\documentclass{article}

\PassOptionsToPackage{numbers, compress}{natbib}

\usepackage[main, final]{neurips_2026}

\usepackage{graphicx}

\usepackage{wrapfig}
\usepackage{subcaption}
\usepackage[utf8]{inputenc} 
\usepackage[T1]{fontenc}    
\usepackage{hyperref}       
\usepackage{url}            
\usepackage{booktabs}       
\usepackage{amsfonts}       
\usepackage{nicefrac}       
\usepackage{microtype}      
\usepackage{xcolor}         
\usepackage{amsmath}
\usepackage{cleveref}
\usepackage{booktabs}
\usepackage{enumitem}
\usepackage{tabularx}
\usepackage{multirow}
\usepackage{makecell}
\usepackage{listings}
\usepackage[table]{xcolor}
\usepackage{lstlinebgrd}
\usepackage{natbib}
\definecolor{parentAcol}{RGB}{215, 232, 255}  
\definecolor{parentBcol}{RGB}{255, 228, 210}  
\definecolor{codekw}{RGB}{0, 70, 150}
\definecolor{codecm}{RGB}{120, 120, 120}
\definecolor{codestr}{RGB}{150, 50, 50}

\lstdefinestyle{pycode}{
    language=Python,
    basicstyle=\ttfamily\scriptsize,
    keywordstyle=\color{codekw}\bfseries,
    commentstyle=\color{codecm}\itshape,
    stringstyle=\color{codestr},
    showstringspaces=false,
    keepspaces=true,
    columns=fullflexible,
    tabsize=4,
    breaklines=true,
    aboveskip=0.3em,
    belowskip=0.3em,
}
\usepackage{array}
\title{Do LLMs Learn from Rewards in Context? : Rethinking the role of reward in In-Context RL}

\author{%
  Minchan Kwon \quad
  Seunghee Koh \quad
  Sunghyun Baek \quad
  Minsung Bae \quad
  Junmo Kim \\
  Korea Advanced Institute of Science and Technology (KAIST) \\
  \texttt{\{kmc0207,seunghee1215,baeksh,ms021230,junmo.kim\}@kaist.ac.kr}
}

\begin{document}

\maketitle

\begin{abstract}

LLM agents increasingly improve at inference time by accumulating experience in context rather than by updating parameters. This process is often described as in-context reinforcement learning (ICRL). Whether in-context learning (ICL) can actually play the role of RL, however, has not been tested. We study this question in its simplest form, direct ICRL, where the model conditions directly on raw trajectory--reward pairs, and ask whether the reward acts as a learning signal.
Through controlled experiments on four benchmarks across six models, we find that the reward is read, but its effect is small: flipping, randomizing, or removing the reward leaves the improvement curve almost unchanged, and this holds even under meta-prompts that explicitly instruct the model to explore, exploit, or reason over rewards. Trajectories drive improvement, but not through their semantic content: shuffled or corrupted trajectories work as well as real ones. These patterns closely mirror those known in ICL, suggesting that direct ICRL is better understood as a special case of ICL than as inference-time RL. This reframing has implications for agent memory design: ICL factors such as input distribution and demonstrations may matter more than RL elements such as reward shaping and exploration.

\end{abstract}
\section{Introduction}

Recently, agent-based AI systems have demonstrated remarkable performance in many areas by learning from experience without changing their parameters~\cite{shinn2023reflexion,zhao2024expel,wang2023voyager,wang2024awm,xia2025experience}.
This involves the agent performing an action, receiving feedback, storing the results in memory, and then acting based on that memory in the next attempt. 
The terminology surrounding these systems is borrowed from reinforcement learning (RL), and the performance improvements they demonstrate are often accepted as evidence of in-context reinforcement learning (ICRL)~\cite{song2025reward,chen2025filtering,shinn2023reflexion}. 
However, in-context learning (ICL) is known to operate quite differently from gradient-based learning~\cite{min2022rethinking,xie2022explanation}. 
ICL is sensitive to input distributions and surface forms, yet surprisingly insensitive to label content~\cite{min2022rethinking}. 
This raises a fundamental question. 
\textit{When an LLM improves through experience-reward pairs in a context, is it performing reinforcement learning, or is it performing ICL?}

To answer this question, we conduct research in the simplest controllable ICRL setting.
We define “Direct In-Context Reinforcement Learning (Direct ICRL).” In this setting, the model is given only raw trajectory–reward pairs. 
This is a simple setting that excludes various additional elements such as memory storage, selection, and summarization methods~\cite{shinn2023reflexion,zhao2024expel,xia2025experience}.
Recent research reports that even in this setting alone, LLMs can function as in-context RL learners~\cite{song2025reward,chen2025filtering}. 
We conduct various controlled experiments in direct ICRL to analyze how direct ICRL works.


\begin{itemize}
    \item[\textbf{(RQ1)}] \emph{The role of reward.} Does the reward text act as a signal that induces an RL-like shift in the policy distribution, boosting high-reward trajectories and suppressing low-reward ones?
    \item[\textbf{(RQ2)}] \emph{The role of trajectory.} Which features of the trajectory drive improvement — its semantic content, its surface form, or its relevance to the task?
\end{itemize}

To answer these questions, we design three controlled experiments.
The first asks whether the improvement depends on the reward.
We keep the trajectories fixed and replace the true rewards with flipped or random values, or remove them entirely (\Cref{sec:exp1}).
The second asks how strongly a reward changes the model's tendency to reproduce a trajectory.
We measure the probability of regenerating each trajectory in memory and how often the model actually replays it (\Cref{sec:exp2}).
The third asks which properties of a trajectory make it useful.
We compare memories built from different sources and perturb trajectories so that their surface form is preserved while their content is corrupted (\Cref{sec:exp3}).
The experiments are conducted across three benchmarks—coding environments (HumanEval+~\cite{chen2021humaneval}, MBPP+~\cite{austin2021mbpp}) and the agentic environment (ScienceWorld~\cite{wang2022scienceworld})—using six models (Qwen-3-4B-Instruct~\cite{qwen3technicalreport}, Qwen-3.5-9B~\cite{qwen3.5}, Llama-3.1-8B~\cite{grattafiori2024llama}, Gemini-3.1-Flash-Lite~\cite{google_gemini31_flash_lite_2026}, GPT-5-nano~\cite{openai_gpt5_nano_2026}, GPT-5-mini~\cite{openai_gpt5_mini_2026}) spanning both open-source and proprietary models.

Our findings indicate that direct ICRL is driven more by trajectory imitation than by reward semantics.
The reward works, but its effect is weak.
Changes in reward levels have a negligible impact on policy probabilities compared to the influence of trajectories.
According to our regeneration-probability analysis, the reward's impact is only about 14\% of the trajectory's.
Furthermore, the model uses trajectories without depending on their semantic correctness. 
Even shuffled or corrupted trajectories perform comparably to real ones.
The only factor that degrades performance is task relevance.
These findings closely mirror the patterns observed in ICL by \citet{min2022rethinking}.

Therefore, we argue that direct ICRL should be reframed as a special case of ICL rather than as inference-time RL. 
This raises questions for agentic memory design, which has so far been guided by RL-style ideas such as reward-based filtering~\cite{chen2025filtering}, exploration~\cite{song2025reward}, and reward shaping. 
The factors that matter may shift toward those identified in ICL, such as input distribution and demonstrations~\cite{min2022rethinking}. 
For example, if failures are accumulated in memory, this may induce replay rather than exploration.
Relatedly, this implies that the success condition for direct ICRL is bounded by the model’s prior.
We discuss these topics in ~\Cref{sec:discussion}.



\section{Related Work}
\label{sec:related}

\paragraph{In-Context Reinforcement Learning.}

In-context reinforcement learning (ICRL) addresses reinforcement learning (RL) problems by adjusting the context.
ICRL methods can be broadly grouped into three categories.
First, some methods require extra training ~\cite{laskin2022context,lee2023supervised,huang2024context,martinez2025context}. These methods introduce additional training methods to acquire ICRL capabilities.
They train models to adapt quickly in specific environments or improve pattern-matching abilities.
Second, abstraction-based methods~\cite{shinn2023reflexion,zhao2024expel,xia2025experience} use additional LLM forward passes or other functions to transform trajectory--reward pairs into natural-language instructions.
Reflexion~\cite{shinn2023reflexion} uses a method of writing reflective sentences, while Expel~\cite{zhao2024expel} constructs them in insight form.
Finally, some methods allow the model to find better trajectories by stacking trajectory--reward pairs~\cite{song2025reward,chen2025filtering}. We refer to this category as \textbf{direct ICRL}.
These methods improve performance by adding a search phase or optimizing the memory allocation.
In this paper, we focus on direct ICRL as the variant that most closely resembles RL and examine its underlying mechanisms.



 \paragraph{In-Context Learning.}
In-context learning (ICL) refers to the phenomenon in which a model solves problems based on the context, without requiring additional training.
It is one of the core capabilities of large language models (LLMs) and provides a foundation for chain-of-thought~\cite{wei2022chain}, self-correction~\cite{kumar2024training}, and even Agentic AI~\cite{shinn2023reflexion,zhao2024expel}.
How ICL works has been a major focus of academic interest, and extensive research has been conducted on the topic.
\citet{min2022rethinking} demonstrated that the key factors in ICL are not labels, but rather the input distribution and surface form.
Subsequent work has examined ICL from several perspectives, including attention-level studies~\cite{olsson2022induction} and task-level studies~\cite{xie2022explanation}.
Recently, counterarguments have been reported suggesting that ICL performs well when models are large and have ample memory.~\cite{wei2023larger}
ICRL is closely related to ICL, but few studies analyze the two together.
To our knowledge, this is the first study to analyze ICRL within the ICL framework, bridging these two lines of work.

\begin{figure}[t]
\centering
\includegraphics[width=1.0\linewidth]{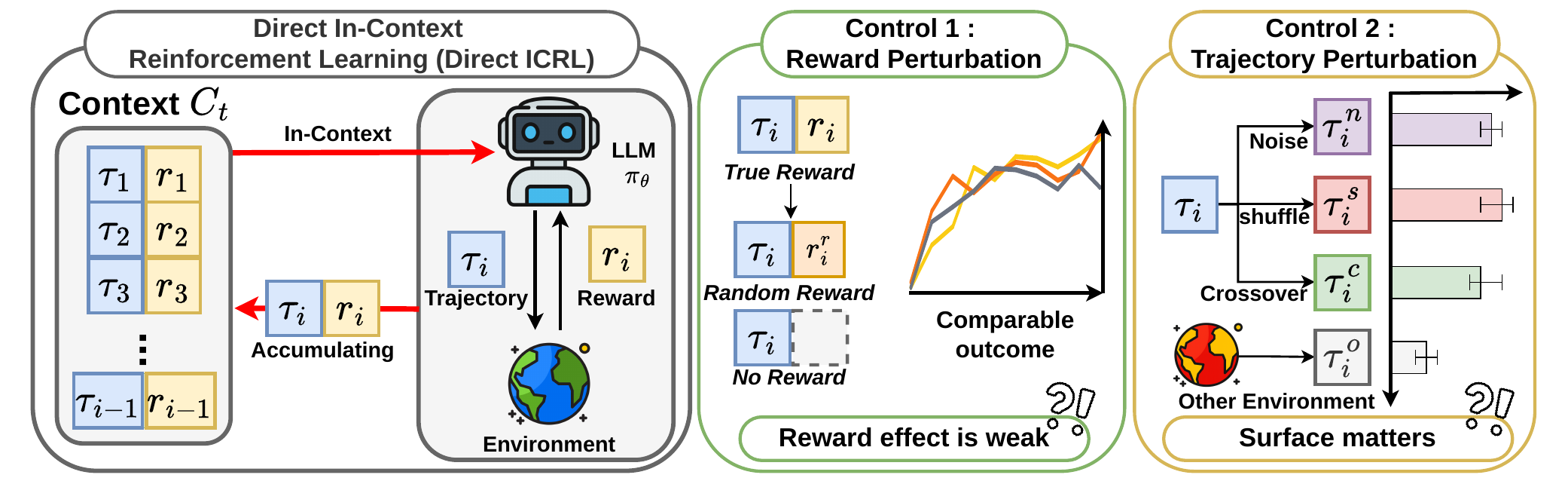}
\caption{Direct ICRL overview (left) and controlled interventions for probing its mechanism: reward (middle) and trajectory perturbations (right).}
\label{fig:Intro}
\end{figure}
\section{Preliminaries}

\subsection{Reinforcement Learning}

We consider a Markov Decision Process (MDP) $(\mathcal{S}, \mathcal{A}, P, R, \gamma)$, where an agent follows a policy $\pi_\theta(a \mid s)$ to produce a trajectory $\tau = (s_0, a_0, s_1, \ldots)$ with return $R(\tau) = \sum_t \gamma^t r_t$. RL maximizes the expected return
\begin{equation}
J(\theta) = \mathbb{E}_{\tau \sim \pi_\theta}[R(\tau)]
\end{equation}
by updating parameters $\theta$.
In this formulation, $\mathcal{S}$ and $\mathcal{A}$ denote the state and action spaces, respectively.
The transition kernel $P(s' \mid s,a)$ defines the probability of reaching state $s'$ after taking action $a$ in state $s$, and $R$ denotes the reward function that provides the scalar reward $r_t$ at time step $t$.
The discount factor $\gamma \in [0,1]$ determines the contribution of future rewards to the return.
The policy $\pi_\theta(a \mid s)$ is a parameterized mapping from states to action probabilities, where $\theta$ denotes the policy parameters.
A trajectory $\tau$ is generated by repeatedly sampling actions from $\pi_\theta$ and states from the transition dynamics $P$.
The objective $J(\theta)$ is the expected discounted return, and RL seeks to find parameters $\theta$ that maximize this objective.

\subsection{In-Context Reinforcement Learning}

In-context reinforcement learning (ICRL) extends ICL to RL problems by adapting a policy through contextual experience rather than parameter updates. Given a task description $q$ and a frozen language model $\pi_\theta$, the model's behavior is shaped by a context $C_t$ for episode $t$ that accumulates information from past interactions:
\begin{equation}
C_t = \big(\phi(\tau_1, r_1),\ \phi(\tau_2, r_2),\ \ldots ,\phi(\tau_{t-1}, r_{t-1})\big),
\end{equation}
where each past trajectory $\tau_i$ and its reward $r_i$ are encoded by a rendering function $\phi$ into a segment of the context. The induced policy $\pi_\theta(\cdot \mid q, C_t)$ thus changes with $t$ even though $\theta$ is fixed. Different instantiations of ICRL differ primarily in the choice of $\phi$ --- what information is retained from each interaction and in what form.

\paragraph{Direct ICRL.} 
We focus on cases where $\phi$ represents each trajectory--reward pair directly, without any additional abstraction or summarization. 
Specifically, $\phi(\tau_i, r_i)$ serializes $\tau_i$ and $r_i$ into text using a fixed template (e.g., \texttt{Trajectory: [$\tau_i$]\textbackslash nReward: [$r_i$]}). 
The resulting $C_t$ is a concatenation of the raw trajectory--reward pairs. We refer to this case as \emph{direct ICRL}.
We focus our analysis on direct ICRL; unless otherwise specified, ICRL refers to direct ICRL for convenience.
A visual overview is provided in~\Cref{fig:Intro}.

\subsection{Tasks}
We evaluate direct ICRL in two regimes.
\textit{Coding} tasks treat each task as a single-action MDP, a format common in LLM-based RL and preference-optimization settings such as DPO~\cite{rafailov2023direct}.
This setting lets us intervene precisely on reward signals while holding trajectory content fixed, reducing confounds from self-bootstrapping and making similarity-based behavior easier to inspect.
\textit{Agentic} tasks involve multi-step interaction, where reward decomposes across sub-goals, and credit assignment across actions is structurally meaningful.
As a closer analogue of standard RL settings, this regime tests whether direct ICRL can handle sequential decision-making and action-level credit assignment.

\paragraph{Coding.} MBPP+~\cite{austin2021mbpp} and HumanEval+~\cite{chen2021humaneval} are instantiated as single-action MDPs: the task description $q$ is the initial state, the action $a_0$ is a complete program emitted by $\pi_\theta$, the episode terminates upon emission, and the reward $R(\tau) \in [0,1]$ is the unit-test pass rate. A trajectory reduces to $\tau = (q, a_0)$. Where $q$ is the task description. 

\paragraph{Agentic.} ScienceWorld~\cite{wang2022scienceworld} is a text-based environment simulating an elementary science curriculum, in which an agent interacts with the environment over multiple steps by issuing high-level text actions (e.g., \texttt{move to kitchen}, \texttt{focus on metal fork}) until the episode terminates.
At each step, the agent observes the current textual state, selects an action according to $\pi_\theta$, and receives a reward from the environment.
Consequently, each ScienceWorld episode corresponds to a multi-step trajectory $\tau = (s_0,a_0,s_1,a_1,\ldots,s_T)$, rather than the single-action trajectory used for coding tasks.
Where $s_t,a_t$ are the state and action at time $t$.
We sample $40$ ScienceWorld tasks uniformly across environment types.
Details of the reward and score settings are provided in~\Cref{appen:SW_reward}.
\section{Reward Semantics Do Not Drive Improvement}
\label{sec:exp1}
\begin{figure}[t]
\centering
\begin{subfigure}{0.32\linewidth}
    \centering
    \includegraphics[width=\linewidth]{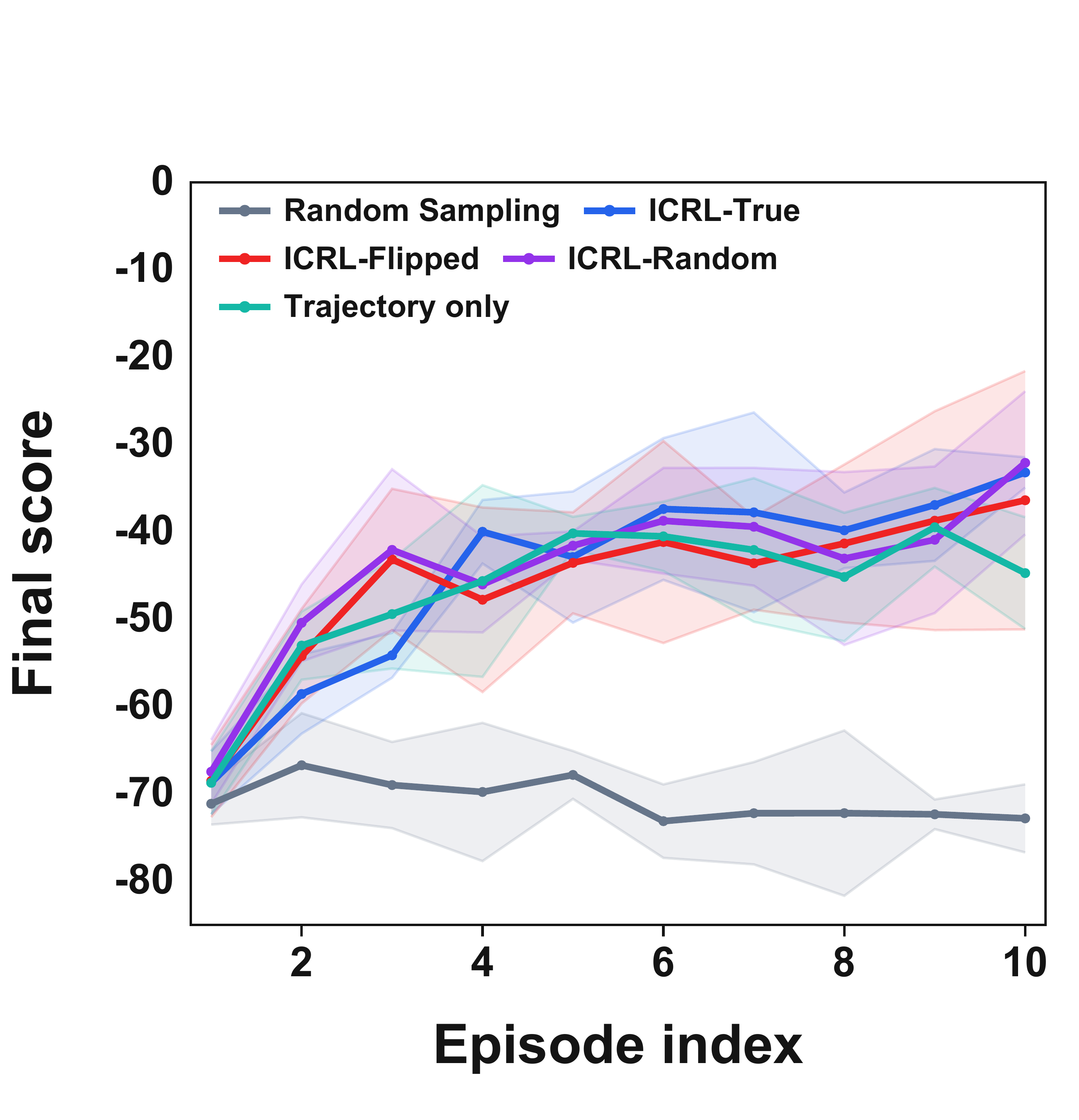}
    \caption{ScienceWorld}
    \label{fig:subfig3}
\end{subfigure}
\hfill
\begin{subfigure}{0.32\linewidth}
    \centering
    \includegraphics[width=\linewidth]{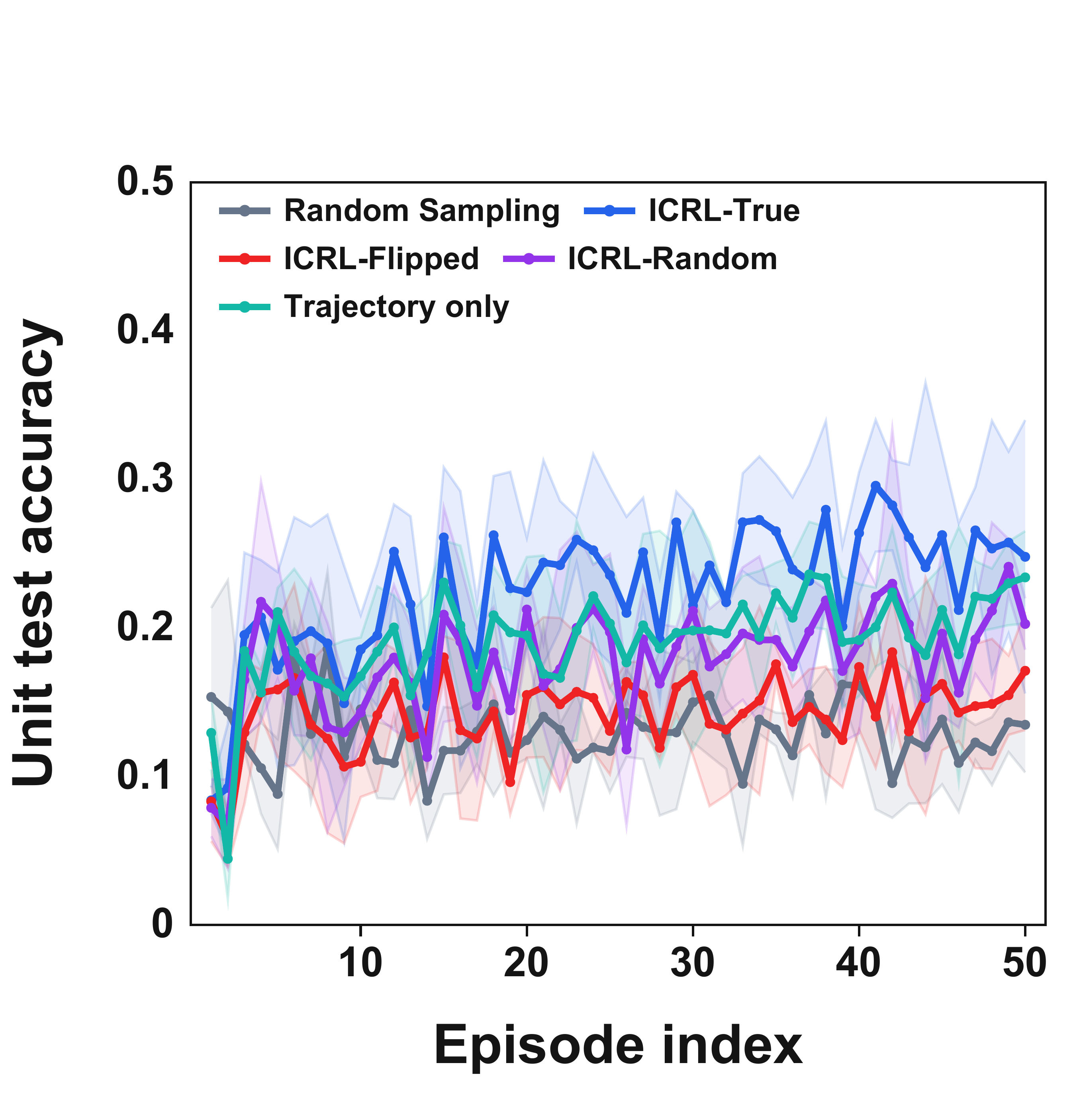}
    \caption{MBPP+}
    \label{fig:subfig2}
\end{subfigure}
\hfill
\begin{subfigure}{0.32\linewidth}
    \centering
    \includegraphics[width=\linewidth]{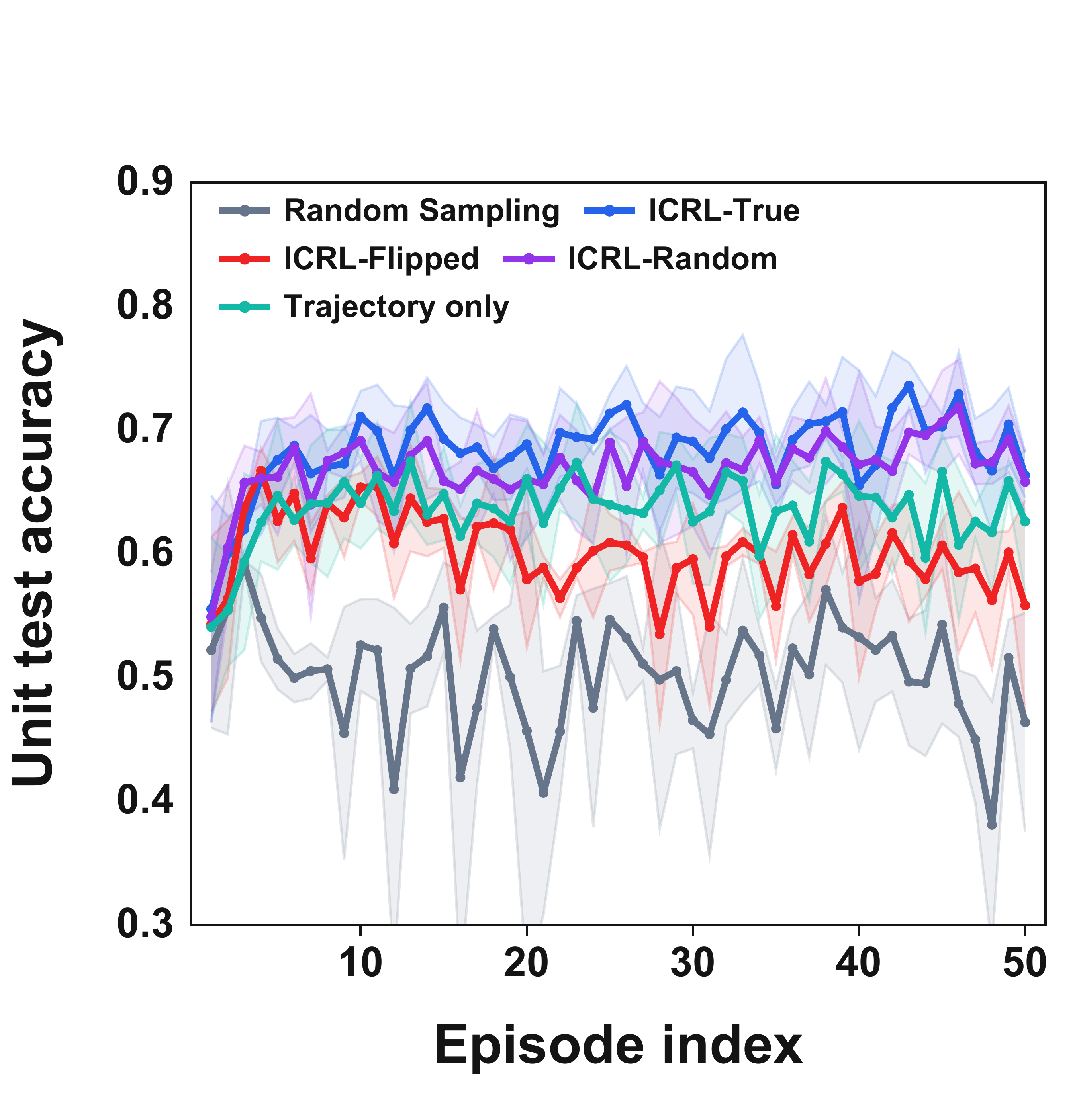}
    \caption{HumanEval+}
    \label{fig:subfig1}
\end{subfigure}

\caption{Episode-level performance under different reward conditions on three benchmarks.}
\label{fig:Acculumate_ICRL_curve}
\vspace{-5pt}
\end{figure}

This section addresses RQ1: Does the reward field act as an RL-aligned signal?
To answer this question, we intervene in the reward field while holding trajectory content fixed: true rewards are replaced with flipped, randomized, or removed values, with trajectories left unchanged.
We track this in the episode-over-episode trajectory of improvement (\Cref{sec:accumulating}) and in a memory-fixed control that removes the trajectory-composition confound (\Cref{sec:fixed_memory}).
Two complementary analyses reached the same conclusion: while direct ICRL outperforms the random sampling baseline, this performance improvement is largely unaffected by the content of the reward. 

\subsection{Effect of Reward in the Memory Accumulation Setting}
\label{sec:accumulating}

\subsubsection{Setup} Each task begins with an empty memory $C_0 = \emptyset$.
At each episode $t$, the model generates a trajectory $\tau_t$ conditioned on $(q, C_{t-1})$ and receives a reward $r_t$.
The trajectory--reward pair $(\tau_t, r_t)$ is appended to the memory sequence as $C_t = C_{t-1} \oplus (\tau_t, r_t)$ where $\oplus$ denotes sequence concatenation.
We compare four reward conditions, identical in trajectory content but differing only in the reward field: \textbf{ICRL-True} ($r_t = R(\tau_t)$), \textbf{ICRL-Flipped} ($r_t = r_{\max} - R(\tau_t) + r_{min}$), \textbf{ICRL-Random} ($r_t \sim \text{Unif}(\mathcal{R})$, sampled from the empirical reward distribution $\mathcal{R}$), and \textbf{Trajectory Only} (the reward field is omitted). For non-ICRL baseline, we use \textbf{Random Sampling (RS)}, which generates without any context ($C=\emptyset$), serves as the no-context baseline.
We run for $10$ episodes on ScienceWorld and $50$ episodes on HumanEval+ and MBPP+, all with $3$ seeds. When $|C_t|$ exceeds the prompt budget of $10$, prompts are sampled uniformly at random from $C_t$.
We select 40 random tasks from each benchmark.
We use Qwen-3.5-9B as the policy model.

\subsubsection{Results}
\paragraph{Direct ICRL improves over Random Sampling.} 
~\Cref{fig:Acculumate_ICRL_curve} shows the mean and standard deviation of episode reward across the three benchmarks. 
ICRL-True improves over RS by a clear margin on every benchmark. This reproduces the standard direct ICRL gain and fixes the effect we want to explain.

\paragraph{Even wrong rewards do not break direct ICRL.} 
ICRL-Flipped is the strongest intervention in this section: it inverts the reward signal entirely, so that high-reward trajectories are labeled as low-reward, and vice versa.
Under an RL view, performance should collapse to or below RS. 
However, ICRL-Flipped remains well above RS on every benchmark, and its curve is closer to ICRL-True than to RS. 
The reward is read (Flipped does sit slightly below True on HumanEval and MBPP), but the gap is small.
Even a full sign flip is not enough to change the direction of direct ICRL.

\paragraph{Random or omitted rewards remain competitive with true rewards.} ICRL-Random matches ICRL-True within seed variance on every benchmark. Trajectory Only, which removes the reward field entirely, also tracks ICRL-True closely on every benchmark and produces the same episode-over-episode improvement. Replacing real reward with random values, or dropping the reward field altogether, leaves the improvement curve essentially unchanged --- the gain from direct ICRL requires neither correctly valued rewards nor a reward field at all.
In summary, while the rewards have an effect, they do not significantly alter the shape of the curve.
This demonstrates that the behavior of direct ICRL differs from that expected in RL, providing motivation for further analysis.


\subsection{Effect of Reward in the Fixed Memory Setting}
\label{sec:fixed_memory}
\begin{table}[t]
\centering
\small
\setlength{\tabcolsep}{3.5pt}
\renewcommand{\arraystretch}{1.08}

\caption{
Comparison of improvement types across models on HumanEval+ and MBPP+.
Each entry reports the number of problems and the percentage within each model-benchmark setting.
}
\label{tab:improvement_by_benchmark}

\begin{minipage}{\linewidth}
\centering
(a) HumanEval+

\vspace{0.35em}

\begin{tabularx}{\linewidth}{
>{\raggedright\arraybackslash}X
>{\centering\arraybackslash}p{0.17\linewidth}
>{\centering\arraybackslash}p{0.20\linewidth}
>{\centering\arraybackslash}p{0.21\linewidth}
>{\centering\arraybackslash}p{0.07\linewidth}
}
\toprule
{Model}
& {{No Imp.}}
& {{Reward-Agnostic}}
& {{Reward-Sensitive}}
& {Total} \\
\midrule
Qwen-3-4B-Instruct    & 25 (83.3\%) & 4 (13.3\%)  & 1 (3.3\%) & 30 \\
Llama-3.1-8B-Instruct & 27 (90.0\%) & 2 (6.7\%)   & 1 (3.3\%) & 30 \\
Qwen-3.5-9B           & 19 (63.3\%) & 10 (33.3\%) & 1 (3.3\%) & 30 \\
GPT-5-nano            & 11 (36.7\%) & 19 (63.3\%) & 0 (0.0\%) & 30 \\
GPT-5-mini            & 7 (23.3\%)  & 22 (73.3\%) & 1 (3.3\%) & 30 \\
Gemini-3.1-Flash-Lite & 12 (40.0\%) & 16 (53.3\%) & 2 (6.7\%) & 30 \\
\midrule
\textbf{Total}
& \textbf{101 (56.1\%)}
& \textbf{73 (40.6\%)}
& \textbf{6 (3.3\%)}
& \textbf{180} \\
\bottomrule
\end{tabularx}
\end{minipage}

\vspace{1.1em}

\begin{minipage}{\linewidth}
\centering
{(b) MBPP+}

\vspace{0.35em}

\begin{tabularx}{\linewidth}{
>{\raggedright\arraybackslash}X
>{\centering\arraybackslash}p{0.17\linewidth}
>{\centering\arraybackslash}p{0.20\linewidth}
>{\centering\arraybackslash}p{0.21\linewidth}
>{\centering\arraybackslash}p{0.07\linewidth}
}
\toprule
{Model}
& {{No Imp.}}
& {{Reward-Agnostic}}
& {{Reward-Sensitive}}
& {Total} \\
\midrule
Qwen-3-4B-Instruct    & 25 (83.3\%) & 4 (13.3\%)  & 1 (3.3\%) & 30 \\
Llama-3.1-8B-Instruct & 23 (76.7\%) & 6 (20.0\%)  & 1 (3.3\%) & 30 \\
Qwen-3.5-9B           & 22 (73.3\%) & 7 (23.3\%)  & 1 (3.3\%) & 30 \\
GPT-5-nano            & 12 (40.0\%) & 16 (53.3\%) & 2 (6.7\%) & 30 \\
GPT-5-mini            & 14 (46.7\%) & 15 (50.0\%) & 1 (3.3\%) & 30 \\
Gemini-3.1-Flash-Lite & 16 (53.3\%) & 12 (40.0\%) & 2 (6.7\%) & 30 \\
\midrule
\textbf{Total}
& \textbf{112 (62.2\%)}
& \textbf{60 (33.3\%)}
& \textbf{8 (4.4\%)}
& \textbf{180} \\
\bottomrule
\label{tab:exp1_fixed_categories}
\end{tabularx}
\end{minipage}
\vspace{-15pt}
\end{table}

We next use a fixed-memory control to isolate the effect of reward.
In \Cref{sec:accumulating}, memory accumulates across episodes, so the trajectories in memory differ across reward conditions.
Here, memory is not accumulated, and all reward conditions share the same trajectories, differing only in the reward field.

\subsubsection{Setup} 
We fix the memory $C$ across all conditions. For each task, we draw five trajectories via random sampling and use them as the fixed prompt across all reward conditions. We average five evaluation samples for each task and memory $C$, and then average the result over three seeds to obtain the final per-task score. 
We use the same baselines in \Cref{sec:accumulating}.
We use the three improvement categories:
\begin{itemize}[leftmargin=1.5em]
    \item \textbf{No Improvement (No Imp.)}: ICRL-True $<$ RS. No improvement occurs.
    \item \textbf{Reward-Agnostic Improvement (Reward-Agnostic)}: ICRL-True $\leq$ (ICRL-Random \textbf{or} ICRL-Flipped) and ICRL-True $>$ RS. 
    \item \textbf{Reward-Sensitive Improvement (Reward-Sensitive)}: ICRL-True $>$ (ICRL-Random \textbf{and} ICRL-Flipped) and ICRL-True $>$ RS. 
    
\end{itemize}

We evaluate six models spanning open-weights and proprietary families: Qwen-3-4B-Instruct~\cite{qwen3technicalreport}, Qwen-3.5-9B~\cite{qwen3.5}, Llama-3.1-8B~\cite{grattafiori2024llama}, Gemini-3.1-Flash-Lite~\cite{google_gemini31_flash_lite_2026}, GPT-5-nano~\cite{openai_gpt5_nano_2026}, and GPT-5-mini~\cite{openai_gpt5_mini_2026}. Experiments are run on HumanEval and MBPP. Detailed hyperparameters by model are provided in Appendix~\ref{appen:Model-Setting}.

\subsubsection{Results}
\Cref{tab:exp1_fixed_categories} shows the results.
Of the 360 task-model pairs, only 14 (3.9\%) were classified as reward-sensitive, 133 (36.9\%) were reward-agnostic, and the remaining 213 (59.2\%) showed no improvement in performance over random sampling. The number of reward-sensitive tasks per model remained between 0 and 2 out of 30 across all six models, spanning open-source 4B--9B systems and proprietary mid-range systems. 
In other words, when memory is fixed, the direct influence of reward content is small.

\section{The Effect of Trajectory Exposure is Stronger than Rewards}
\label{sec:exp2}

\Cref{sec:exp1} shows that reward content has little effect on task performance.
This section asks which part of memory is important by examining what the model does with each trajectory in its memory.
The two views of direct ICRL make different predictions.
If direct ICRL works like RL, the reward attached to a trajectory should control how likely the model is to produce that trajectory again: high-reward trajectories should be replayed more often, and low-reward trajectories less often.
If it works like ICL, showing a trajectory should make it more likely to be replayed, whatever its reward.
We test these predictions with three questions: whether showing a trajectory increases its replay (\Cref{sec:5.2}), whether the reward changes that replay (\Cref{sec:5.3}), and what else changes it (\Cref{sec:5.4}).

\subsection{Setup}
\label{sec:5.1}

\paragraph{Memory and targets.}
For each task, we build a fixed memory $\mathcal{M} = \{(\tau_i, r_i)\}_{i=1}^{N}$ of $N=10$ trajectory--reward pairs by random sampling on HumanEval and MBPP.
We track two target trajectories $\tau^* \in \mathcal{M}$ per task: $\tau^-$, the lowest-reward entry, and $\tau^+$, the highest-reward entry.
We use Qwen-3.5-9B as the policy model.

\paragraph{Metrics.}
We measure replay at the probability level and at the behavior level.
At the probability level, we use the leave-one-out difference
\begin{equation}
\Delta\mathrm{LOO}(\tau^*) = \log \pi_\theta(\tau^* \mid q, m, \mathcal{M}) - \log \pi_\theta(\tau^* \mid q, m, \mathcal{M} \setminus \{\tau^*\}),
\end{equation}
which measures how much more likely $\tau^*$ becomes simply because it is present in the memory.
Here $m$ is the meta-prompt (Appendix~\ref{appen:Meta-Prompt}), and all log-probabilities are averaged per token.

At the behavior level, we sample 500 generations per condition.
\emph{Target Replay Rate} is the fraction of samples whose normalized abstract syntax tree (AST) matches $\tau^*$.
\emph{Memory Replay Rate} is the fraction matching any trajectory in $\mathcal{M}$.

Each question below changes one aspect of the memory, and we introduce the corresponding conditions where they are used.

\subsection{Showing a Trajectory Makes the Model Replay It}
\label{sec:5.2}

We first compare two conditions.
\emph{Problem-only} gives the model only the task description $q$, with no memory.
\emph{Trajectory-only} adds the trajectories in $\mathcal{M}$, without the reward field.

\begin{table}[t]
\centering
\caption{Effect of showing trajectories, without any reward. Replay rates are in \%. $\Delta\mathrm{LOO}$ is undefined for Problem-only because no memory is shown.}
\label{tab:exposure}
\begin{tabular}{lccccc}
\toprule
 & \multicolumn{2}{c}{$\Delta\mathrm{LOO}$} & \multicolumn{2}{c}{Target Replay Rate} & Memory Replay Rate \\
\cmidrule(lr){2-3} \cmidrule(lr){4-5}
Condition & $\tau^-$ & $\tau^+$ & $\tau^-$ & $\tau^+$ & \\
\midrule
Problem-only    & --    & --    & 0.7 & 1.1 & 16.3 \\
Trajectory-only & 0.416 & 0.162 & 3.1 & 1.8 & 34.3 \\
\bottomrule
\end{tabular}
\end{table}

\Cref{tab:exposure} shows the result.
Under Problem-only, 16.3\% of generations happen to match a trajectory in $\mathcal{M}$; this is the rate at which the model produces these outputs on its own.
Showing the trajectories raises the Memory Replay Rate to 34.3\%, more than twice that level.
The same holds for individual targets: the Target Replay Rate rises from 0.7\% to 3.1\% for $\tau^-$ and from 1.1\% to 1.8\% for $\tau^+$, and $\Delta\mathrm{LOO}$ is positive for both (0.416 and 0.162).
The increase is larger for the lowest-reward trajectory than for the highest-reward one, even though no reward is shown.
Exposure alone is therefore enough to induce replay, and it does not favor good trajectories.
We use the $\tau^-$ value of 0.416 as the reference size of this trajectory-presence effect.

\subsection{Reward Shifts Replay in the RL Direction, but Only Slightly}
\label{sec:5.3}

We next add the reward field to the same memory.
\emph{ICRL} attaches the true reward to every trajectory.
\emph{Specific-flip} is identical to ICRL except that the reward of the target trajectory is flipped ($r_{\max} - r^* + r_{\min}$), so that $\tau^-$ is labeled as the best trajectory and $\tau^+$ as the worst.
The RL view predicts that true rewards lower the replay of $\tau^-$ and raise that of $\tau^+$, and that flipping the target's reward reverses both.

\begin{table}[t]
\centering
\caption{Effect of the reward field on the replay of the target trajectory. Trajectories are identical across rows; only the reward field differs. Target Replay Rate is in \%.}
\label{tab:replay-attractor}
\begin{tabular}{lcccc}
\toprule
 & \multicolumn{2}{c}{$\Delta\mathrm{LOO}$} & \multicolumn{2}{c}{Target Replay Rate} \\
\cmidrule(lr){2-3} \cmidrule(lr){4-5}
Condition & $\tau^-$ & $\tau^+$ & $\tau^-$ & $\tau^+$ \\
\midrule
Trajectory-only (no reward) & 0.416 & 0.162 & 3.1 & 1.8 \\
ICRL (true reward)          & 0.360 & 0.173 & 2.5 & 3.0 \\
Specific-flip               & 0.474 & 0.145 & 3.0 & 1.6 \\
\bottomrule
\end{tabular}
\end{table}

\paragraph{The direction matches RL.}
\Cref{tab:replay-attractor} shows that every shift goes in the predicted direction.
With true rewards, $\Delta\mathrm{LOO}$ falls for $\tau^-$ and rises for $\tau^+$, and the Target Replay Rate moves the same way ($-0.6$\,pp for $\tau^-$ and $+1.2$\,pp for $\tau^+$).
Flipping the target's reward reverses both: relative to ICRL, $\Delta\mathrm{LOO}$ rises for $\tau^-$ and falls for $\tau^+$, and the Target Replay Rate changes by $+0.5$\,pp and $-1.4$\,pp.
The model therefore reads the reward and responds to it as RL would.

\paragraph{The size is small.}
Relative to Trajectory-only, attaching a reward changes $\Delta\mathrm{LOO}$ by at most 0.058 ($\tau^-$ under Specific-flip).
This is about 14\% of the 0.416 trajectory-presence effect from \Cref{sec:5.2}.
The behavior-level result agrees: even with its true low reward shown, $\tau^-$ is replayed in 2.5\% of generations, well above the 0.7\% under Problem-only.
A low reward reduces the replay of a failed trajectory, but it does not undo the replay caused by showing that trajectory.

\subsection{Position and Raw Trajectories Change Replay; Reward Labels and Instructions Do Not}
\label{sec:5.4}

If reward has only a small effect, what does change replay?
Starting from ICRL, we apply three kinds of changes.
(i)~\emph{Position}: \emph{Order-first} and \emph{Order-last} move the target pair to the beginning or the end of the memory, leaving its content and reward unchanged.
(ii)~\emph{Discouraging replay}: \emph{All Failure} sets every reward in the memory to 0, and \emph{Anti Prompt} adds an explicit instruction not to repeat the shown trajectories (Appendix~\ref{appen:Meta-Prompt}).
(iii)~\emph{Removing same-task raw trajectories}: \emph{Cross-task} fills the memory with trajectories from a different task, and \emph{Reflexion}~\cite{shinn2023reflexion} replaces the trajectory--reward pairs with natural-language summaries.

\begin{table}[t]
\centering
\caption{Effect of changes other than the target's reward. All replay rates are in \%. The $\tau^-$ and $\tau^+$ columns report runs in which the target is the lowest- and the highest-reward trajectory, respectively. Problem-only is the level without any memory.}
\label{tab:behavior}
\begin{tabular}{llcccc}
\toprule
 & & \multicolumn{2}{c}{Target Replay Rate} & \multicolumn{2}{c}{Memory Replay Rate} \\
\cmidrule(lr){3-4} \cmidrule(lr){5-6}
Change & Condition & $\tau^-$ & $\tau^+$ & $\tau^-$ & $\tau^+$ \\
\midrule
Reference & Problem-only & 0.7 & 1.1 & 16.3 & 16.3 \\
          & ICRL         & 2.5 & 3.0 & 30.5 & 30.5 \\
\midrule
Position & Order-first & 2.4 & 5.5  & 29.3 & 30.0 \\
         & Order-last  & 3.2 & 13.2 & 29.4 & 41.9 \\
\midrule
Discouraging replay & All Failure & 2.3 & 1.7 & 27.6 & 27.3 \\
                    & Anti Prompt & 2.4 & 2.3 & 30.2 & 30.0 \\
\midrule
No same-task raw trajectory & Cross-task & 4.7 & 0.0 & 13.4 & 13.3 \\
                            & Reflexion  & 0.5 & 1.9 & 11.4 & 11.7 \\
\bottomrule
\end{tabular}
\end{table}

\paragraph{Position has a larger effect than reward.}
Moving the target to the end of the memory raises the $\tau^+$ Target Replay Rate from 3.0\% to 13.2\%, and moving it to the beginning raises it to 5.5\% (\Cref{tab:behavior}).
The trajectory and its reward are unchanged; only the position differs.
The largest reward-induced change in \Cref{sec:5.3} was 1.4\,pp, so position alone moves replay several times more than the reward does.
Placing $\tau^+$ last also raises the Memory Replay Rate from 30.5\% to 41.9\%.
This effect is concentrated on $\tau^+$: the $\tau^-$ rate stays within one point of ICRL under both orderings.

\paragraph{Zero rewards and explicit instructions barely reduce replay.}
Labeling every trajectory as a failure lowers the Memory Replay Rate only from 30.5\% to about 27\%.
Instructing the model to try a different direction leaves it at about 30\%.
Both remain far above the 16.3\% of Problem-only.
The lowest-reward target shows the same pattern: $\tau^-$ is replayed at 2.3\% under All Failure and 2.4\% under Anti Prompt, against 2.5\% under ICRL.
Neither the reward signal nor a direct instruction is enough to pull the model away from the trajectories it has been shown.

\paragraph{Removing same-task raw trajectories removes the replay.}
Under Cross-task and Reflexion, the Memory Replay Rate falls to about 13\% and 12\%, below the Problem-only level.
Replay is thus tied to the presence of raw trajectories from the same task, not to the reward information: Reflexion still conveys which attempts failed, yet it does not induce replay.

\paragraph{Summary.}
Showing a trajectory increases its replay, the reward changes this by a small amount in the RL-aligned direction, and position and raw-trajectory presence change it far more.
This matches the pattern known from ICL, where the presence and arrangement of demonstrations matter more than the content of their labels~\cite{min2022rethinking}. 
\section{Surface Form and Task Relevance Drive Improvement}
\label{sec:exp3}

\Cref{sec:exp2} established that direct ICRL improves through trajectory imitation, with reward modulation playing at most a marginal role. 
This sharpens RQ2: if imitation is the operative mechanism, which features of the trajectory make it a useful imitation target — its content, its surface form, or its relevance to the task?
We conduct a controlled memory-composition analysis that varies memory contents while keeping the imitation mechanism fixed, tracing the gain to two properties---\emph{task-relevant distribution} and \emph{surface form}.
We use Qwen-3.5-9B as the policy model.

\subsection{Setup}
\label{sec:exp3_setup}


We construct memories of varying composition and measure the resulting episode reward at fixed model and decoding settings, the same as in \Cref{sec:fixed_memory}.
The only difference is that we set the memory size to $N = 10$ trajectories.
We report the mean reward and standard deviation over five random samples.
We compare eight memory configurations under two categories: source variations and trajectory perturbations.

\textbf{Source variations.}
\begin{itemize}[leftmargin=1.5em]\setlength\itemsep{0pt}
    \item \textbf{RSM (Random Sampling Memory)}: trajectories sampled from $\pi_\theta$ on the same task $q$.
    \item \textbf{ICRL History}: trajectories from the final memory of a full ICRL run (\Cref{sec:exp1}).
    \item \textbf{Cross-Task}: trajectories sampled from a different task $q' \neq q$.
    \item \textbf{Random Valid Action}: arbitrary valid actions; random code-path tokens in MBPP.
\end{itemize}

\textbf{Trajectory perturbations.} These perturbations are applied to the ICRL History.
\begin{itemize}[leftmargin=1.5em]\setlength\itemsep{0pt}
    \item \textbf{Shuffle}: trajectories with reordered its middle parts.
    \item \textbf{Crossover}: trajectories alternating segments from different tasks.
    \item \textbf{Insert Token} (MBPP only): trajectories with a random token inserted to induce a syntax error.
\end{itemize}


\begin{table}[]
\caption{Mean reward across MBPP and ScienceWorld. Standard deviations are reported in ~\Cref{tab:full_6}.}
\centering

\setlength{\tabcolsep}{16pt} 
\begin{tabular}{lcc}
\toprule
Memory Setting & MBPP & ScienceWorld \\ \midrule
RSM & 0.250 & -32.479 \\ \midrule
Cross-Task & 0.122 & -43.000 \\
Random Valid Action & 0.102 & -68.208 \\ \midrule
ICRL History & 0.245 & -42.238 \\ \midrule
\multicolumn{3}{l}{\textit{Trajectory Perturbation}} \\
\quad Shuffle & 0.243 & -38.588 \\
\quad Crossover & 0.251 & -39.701 \\
\quad Insert token & 0.242 & -- \\ \bottomrule
\end{tabular}%
\label{tab:mean_reward_comparison}
\end{table}

\subsection{Results}
\paragraph{Self-bootstrapping does not help.}
\Cref{tab:mean_reward_comparison} shows the results of fixed-memory settings in ScienceWorld and MBPP.
In both settings, ICRL history performs similarly to or worse than RSM.
This suggests that the self-bootstrapping occurring during the ICRL process does not significantly affect performance.
This contradicts the standard expectation in RL that reward signals guide exploration toward more useful trajectories than random sampling.

\paragraph{Surface form is more important than correctness.} 
Trajectory perturbation techniques target the semantic content of a trajectory while preserving its surface form.
Shuffle and Crossover destroy ordering (lines in MBPP, action-state pair in ScienceWorld), and Insert Token additionally breaks compilation. 
If the model relied on this semantic content, all three perturbations should degrade performance.
However, performance under perturbed trajectories is comparable to ICRL History. 
We interpret this as the model relying on the surface form of the trajectory, not its semantic content. 
This trend is most clearly observed in ScienceWorld.
In sequential decision-making, shuffling the order of actions should be fatal, as it breaks the causal chain linking one state to the next.
Nevertheless, the fact that performance is maintained suggests that the model does not track trajectories step-by-step but gains an advantage by simply recognizing the correct action, regardless of its order.

\paragraph{Task relevance is essential}
The improvement requires trajectories that are \emph{actual attempts at the target task}; neither task identity nor surface form alone is enough. 
Cross-Task and Random Valid Action both degrade performance. 
Cross-Task uses real attempts on a different task: its failure shows that matching surface form is not enough — the trajectory must be on the target task. 
(The drop is steeper on MBPP than on ScienceWorld because MBPP code shares little structure across tasks, while ScienceWorld actions and states are largely shared.) 
Random Valid Action uses the target task environment but replaces real attempts with random valid actions or tokens: its dramatic drop shows that being on the right task is also not enough. 
Combined with the perturbation results, the picture is consistent: trajectories need not be structurally correct, but they must be (i) on the target task and (ii) actual attempts at solving it.

\section{Discussion}
\label{sec:discussion}
\paragraph{Direct ICRL is ICL rather than RL.}
The patterns we observed are similar to those identified in ICL. 
\citet{min2022rethinking} demonstrated that ICL remains effective even when the label region is randomized, indicating that form and distribution are more important than content.
Our results extend this perspective to the reward field: direct ICRL rewards operate similarly to ICL labels, and their benefits are driven by the distribution and shape of the trajectories. 
From this perspective, direct ICRL is not a parameter-free RL system, but rather a form of in-context learning that incorporates a reward field.
This suggests that the RL lens through which ICRL has been studied (filtering, self-bootstrapping, reward-based design) needs to be reconsidered. 
ICRL memory design may need to emphasize factors that ICL has identified as central, such as input distribution and surface form.

\paragraph{Failure trajectories provide weak avoidance signals.}
In our experiment, we observed that the negative credit assignment had a negligible effect.
Even in the ICRL-Anti Prompt setting described in~\Cref{tab:behavior}, where the prompt explicitly encouraged the model to ``take a different action,'' the reduction in replay rate was limited.
These results indicate that changing the reward signal alone is insufficient to alter the model's behavior.
Unlike in policy-gradient learning, where repeated failures can indirectly promote exploration in different directions, failed trajectories in direct ICRL may not reliably serve as exploration-inducing signals.
These findings suggest that the common practice of accumulating failed trajectories to drive exploration may not directly apply to direct ICRL.
Abstraction-based approaches that do not rely directly on raw trajectories, such as Reflexion~\cite{shinn2023reflexion} or Expel~\cite{zhao2024expel}, may offer an alternative.
However, stabilizing such methods remains challenging due to their reliance on text gradients~\cite{pryzant2023automatic}.
Developing stable abstraction-based update methods remains a promising direction for future research.

\paragraph{Direct ICRL is bounded by pretrained priors.}
The limited effect of reward semantics on performance suggests two possible interpretations: either the model cannot read the reward content, or it can read it but ignores the reward based on its own preferences regarding which trajectory to mimic. 
Although our experiment did not distinguish between these mechanisms, both point to the same limitation.
The contextual reward semantics did not effectively override the model’s preference for which trajectory to imitate.
Reported successes of direct ICRL are concentrated in environments where high-reward trajectories align with outcomes that a capable baseline model can already generate. 
When the target behavior conflicts with the model’s prior knowledge—such as in new domains, adversarial environments, or exploration-intensive tasks—reward labels alone cannot easily replace explicit memory design or training.
This aligns with previous research findings that negation is difficult in ICL~\cite{vrabcova2025negation}.
These findings raise questions about the scope of ICRL's applicability and, more broadly, whether context alone is enough to adapt to all tasks.

\section{Limitation}
Our experiments are designed to isolate the mechanisms underlying direct ICRL, leaving several boundary conditions outside the scope of our claims.
First, our analysis targets direct ICRL, where raw trajectory--reward pairs are exposed to the model; this setting isolates the roles of trajectory and reward fields but excludes abstraction-based variants such as reflections, skill libraries, or extracted insights.
Second, our results are reported under standard meta-prompts across six models and three benchmarks; we do not rule out prompt formats or model families that elicit stronger reward-aware behavior, including models trained specifically for ICRL, such as the Decision Transformer family~\cite{huang2024context}.
Third, our manipulations probe reward semantics within the studied environments but do not fully characterize counterfactual regimes where reward semantics conflict with the model's prior preferences.

\bibliographystyle{plainnat}
\bibliography{custom}

\newpage
\appendix
\section{Extended Discussion of Limitations}
\label{app:limitations}
Our experiments focus on isolating the mechanisms underlying direct ICRL, rather than exhaustively characterizing all variants and boundary conditions. 
This focus leaves several questions outside the scope of our claims, which we discuss here in more detail.

\paragraph{Scope: direct ICRL only.} 
Our analysis is limited to direct ICRL, where raw trajectory–reward pairs are presented to the model as context. 
This setting matches our primary concern of isolating the roles of the trajectory and reward fields when both are directly exposed to the model.
Some variants transform past experience through abstraction layers, such as natural-language reflections, skill libraries, or extracted insights. 
We exclude them because these transformations introduce additional degrees of freedom that are difficult to control within the same experimental framework. 
Whether our findings extend to such abstraction-based methods remains an open question.

\paragraph{Prompt formats and model families.} 
We observe consistent reward-insensitivity across six models and three benchmarks, but we report results only under standard meta-prompts.
We do not rule out the existence of prompt formats that elicit stronger reward-aware behavior. 
Furthermore, we do not claim that this phenomenon generalizes to all model families. Models trained specifically for ICRL — such as the Decision Transformer family~\cite{huang2024context} — are out of scope and may behave differently, since they are explicitly optimized to read reward signals as control inputs.
Also, our models span 4B–9B open-weights and proprietary mid-range systems; very large (>100B) models remain untested.

\paragraph{Counterfactual reward regimes.} Our experiments test reward manipulations within the studied environments, but do not fully characterize counterfactual reward regimes where reward semantics conflict with the model's prior preferences. 
As discussed in Section~\ref{sec:discussion}, the boundary at which reward signals begin to override pretrained preferences likely depends on the gap between the target behavior and what the base model already prefers. 
Mapping out this boundary — across domains, task types, and reward scales — remains an important direction for future work on direct ICRL.

\section{Experimental Details}
\subsection{Model Setting}
\label{appen:Model-Setting}
We used a temperature of 0.8 for all models except GPT and Gemini. We set the max new tokens to 32 for Scienceworld and to 1024 for MBPP+ and HumanEval+.
We used a thinking intensity of medium for both GPT-5-nano and Mini.
For Gemini, we used a thinking effort of low.

\subsection{ICRL Setting}
\subsubsection{Meta-prompt for each experiment}
\label{appen:Meta-Prompt}

\paragraph{Meta-prompt of HumanEval+ and MBPP+}
"
You are solving the same programming task again.
Below are previous trajectories and their unit test success rates.
Return only the Python code completion for the task prompt that follows.
"

\paragraph{Meta-prompt of ScienceWorld.}
"You are acting in a ScienceWorld task again.
Below are completed previous episodes for this same task and their cumulative returns.
Use those trajectory-and-reward memories to avoid repeating low-return behavior and to imitate higher-return behavior.
When choosing the next action, reflect on which earlier episodes made real progress on the task.
Respond with exactly one valid action and no explanation."

\paragraph{Instruction of Anti-Prompt.}
"
You are revisiting this coding problem after several previous attempts.
Below is a shared memory of earlier trajectories and their rewards.
Do not imitate the repeated failures. Try a meaningfully different direction from the shown history.
Return only Python code.
"

\paragraph{Instruction of Reflexion.}
"You are analyzing previous attempts for one coding problem."
"Task: \{task\_id\}"
"Summarize what is going wrong across the attempts and what a better next attempt should change."
"Write a concise reflection note in bullet-free prose."
"[Problem Prompt]",
prompt.rstrip()

\subsubsection{Experiments Details}
\paragraph{Memory Settings.}
In MBPP+ and HumanEval+, all memories are stored in a memory pool, and 10 are randomly selected from it.
In Top-k and Bottom-k, the top or bottom k samples are retrieved from the memory pool.
In the case of ScienceWorld, all stored memories are used, so there is no memory sampling.

\paragraph{Computational Cost.}
We used one Nvidia A100 80GB for the experiment.
In ~\Cref{fig:Acculumate_ICRL_curve}, ScienceWorld takes about 5 hours for 10 episodes, while HumanEval and MBPP take about 3 hours.

\paragraph{ScienceWorld reward rules.}
\label{appen:SW_reward}
ScienceWorld provides deterministic, environment-defined rewards based on task-specific goal progress.
The simulator maintains a structured world state, updates it through predefined action and science-domain rules, and evaluates progress using task-specific goal monitors.
The Python API reports the simulator score on a 0--100 scale and defines the step reward as the change in score from the previous step.
While most rewards are dense progress signals, catastrophic or task-invalid behavior can drive the score below zero; the API then produces a large penalty such as $-100$.

\section{Additional Experiments}
\subsection{Statistical Analysis of Section 4.1}
\label{appen:static}

\paragraph{Setup.}
We complement Figure~2 with paired $t$-tests on final episode 
scores. For each benchmark, we use 40 tasks and 3 seeds per 
condition. We pair at the task level: for each task, we 
average across seeds to obtain one score per condition, 
yielding 40 paired observations. We prefer this over 
pairing by seed ($n=3$). Seed-level pairing first averages 
across all tasks, which hides the spread between tasks and 
produces misleadingly narrow confidence intervals; task-level 
pairing exposes the actual task-to-task variation.

\paragraph{Results.}
\begin{table}[h]
\centering
\caption{Task-level paired $t$-tests on final episode scores 
($n=40$ tasks per benchmark, paired across 3 seeds). 
All six confidence intervals include zero, and no comparison 
reaches $p<0.05$. ScienceWorld final scores span roughly 80 points; 
coding scores are unit-test accuracy in $[0,1]$.}
\label{tab:stat-confirm}
\begin{tabular}{llrrr}
\toprule
Benchmark & Comparison & Mean $\Delta$ & 95\% CI & $p$ \\
\midrule
\multirow{2}{*}{ScienceWorld} 
  & True vs.\ Random      & $-0.76$  & $[-9.40,\ 7.86]$    & $0.86$ \\
  & True vs.\ No Reward   & $+1.96$  & $[-6.73,\ 10.64]$   & $0.65$ \\
\midrule
\multirow{2}{*}{MBPP+} 
  & True vs.\ Random      & $+0.028$ & $[-0.042,\ 0.098]$  & $0.43$ \\
  & True vs.\ No Reward   & $+0.014$ & $[-0.064,\ 0.091]$  & $0.72$ \\
\midrule
\multirow{2}{*}{HumanEval+} 
  & True vs.\ Random      & $+0.005$ & $[-0.064,\ 0.075]$  & $0.88$ \\
  & True vs.\ No Reward   & $+0.037$ & $[-0.056,\ 0.130]$  & $0.42$ \\
\bottomrule
\end{tabular}
\end{table}
Table~\ref{tab:stat-confirm} reports mean differences, 
95\% confidence intervals, and $p$-values for two comparisons: 
ICRL-True against ICRL-Random and ICRL-True against Trajectory 
Only. Across all three benchmarks, all six confidence 
intervals include zero and all $p$-values exceed $0.4$. 
Mean differences stay below 4\% of each benchmark's score range. 

We read this as a soft \emph{indistinguishable} rather than 
proof of zero effect. The intervals do not exclude small 
positive effects of reward content, but they bound the size 
of any such effect: even at the upper end of each interval, 
the gap remains a small fraction of the total improvement 
ICRL achieves over Random Sampling. This is consistent with 
the per-trajectory analysis in Section~5, which finds a small 
but RL-aligned reward effect that is dominated by the 
trajectory-presence force.

\paragraph{Per-task tests.}
We also run $t$-tests within each task ($n=3$ seeds). With 
this sample size the tests have very low power, and we do 
not treat them as primary evidence. One task (HumanEval/106) 
reached uncorrected $p<0.05$ in the True vs.\ No Reward 
comparison; we treat this as expected sampling noise across 
the $240$ individual tests.

\paragraph{Robustness.}
Wilcoxon signed-rank tests on the same paired data yield 
qualitatively identical conclusions (all $p>0.4$). 
We also examined episode-level tests on ScienceWorld: 
one episode (episode~3) shows uncorrected $p=0.040$ for 
ICRL-True vs.\ ICRL-Random, but this does not survive 
Holm correction across the ten episodes.

\subsection{Per-Task Analysis of Section 4.1}
\label{sec:per-task}

\paragraph{Setup.} To analyze what occurred in each individual task—rather than focusing on the average across the benchmark—we conduct a per-task analysis.
We use the same category in section 4.2.
We perform category classification using the average of the 3 seeds.

\begin{figure}
\centering

\resizebox{0.8\linewidth}{!}{%
    \includegraphics{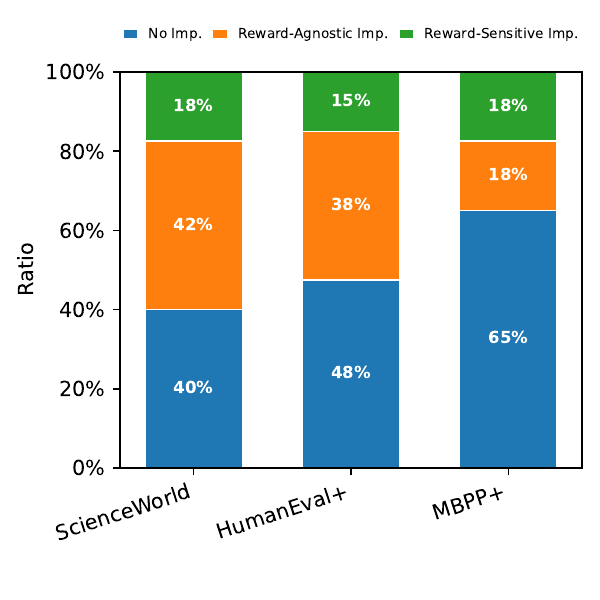}
}

\caption{Improvement category for accumulated memory settings.}
\label{fig:4_1_WrapFigure}
\end{figure}

\paragraph{Results.}
Figure~\ref{fig:4_1_WrapFigure} reports the distribution. 
Surprisingly, “Reward-Agnostic Improvement” accounts for a significant proportion of the improvements across all tasks.
Even MBPP+, which shows the greatest reward-sensitive improvement, accounts for only 50\% of the total improvement.
Scienceworld, which achieves a 42\% reward-agnostic improvement, also exhibits a reward-independent curve in ~\Cref{fig:Acculumate_ICRL_curve}.
This demonstrates that the results from Section 4.1 hold true at the task level as well. While rewards do have an effect, most of the improvement occurs independently of rewards. 

\subsection{Ablation of Section 4.1}

\paragraph{Model Ablation.}
We replicated the Scienceworld experiment described in Section 4.1 using llama-3.1-8B-Instruct.
The experimental results are similar to those discussed in Section 4.1.
True, Flipped, and Random are subject to seed variance.
These results demonstrate that our findings are not specific to Qwen3.5-9B alone.

\paragraph{ScienceWorld Reflexion.} 
We applied Reflexion to the Scienceworld experiment in Section 4.1 of ~\cref{tab:improvement_by_benchmark}.
The results achieved performance comparable to ICRL-True.
This demonstrates the potential of Reflexion. 
Since Reflexion does not use direct trajectories, it is free from the attraction problem.
However, text gradient methods, including Reflexion, are notorious for high variance.
Methods to address this and advance ICRL are left as future work.

\paragraph{ScienceWorld Verbal Reward.}
We conduct the Scienceworld experiment described in Section 4.1 using verbal rewards.
In the verbal reward system, reward scores are divided into four equal parts and categorized into four groups: “failure,” “partial failure,” “partial success,” and “success.”
Verbal rewards also demonstrate performance similar to numerical rewards.
It is worth noting that Flip’s performance is significantly lower than that of numerical rewards.
While Flip and True were indistinguishable in the numerical reward setting, Flip performed worse than True in the verbal reward setting.
This suggests that reward reading ability can be improved through sentence-level enhancements.
However, Flip's performance does not decline monotonically, supporting our claim that the reward effect is masked by attraction.




\begin{figure}[t]
    \centering
    \includegraphics[width=0.7\linewidth]{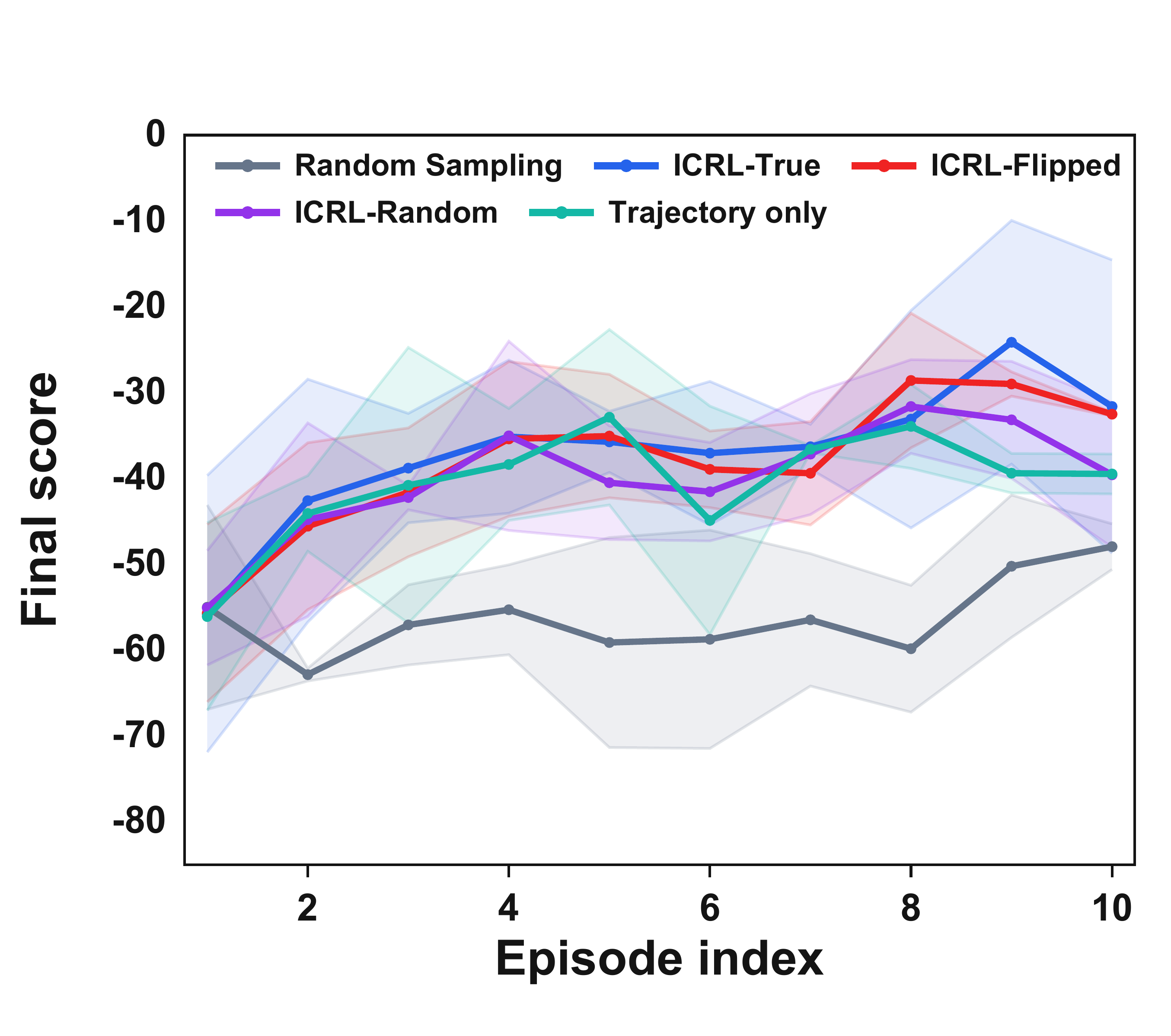}
    \caption{ScienceWorld ICRL experiment using Llama-3.1-8B-Instruct as a policy network.}
    \label{fig:llama_sw}
\end{figure}

\begin{figure}[t]
    \centering
    \includegraphics[width=0.7\linewidth]{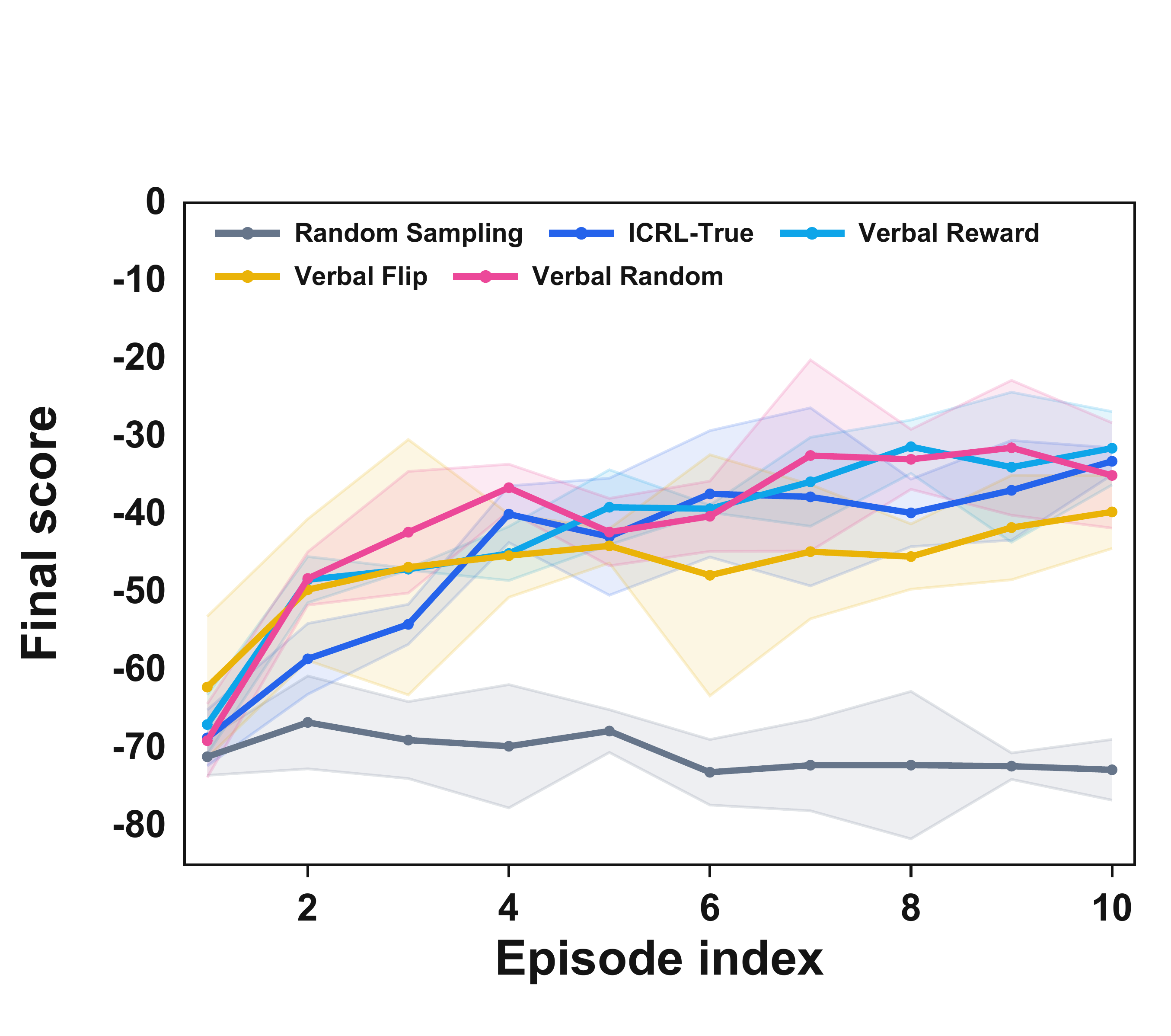}
    \caption{ScienceWorld ICRL experiment with verbal reward.}
    \label{fig:sw_verbal}
\end{figure}

\begin{figure}[t]
    \centering
    \includegraphics[width=0.7\linewidth]{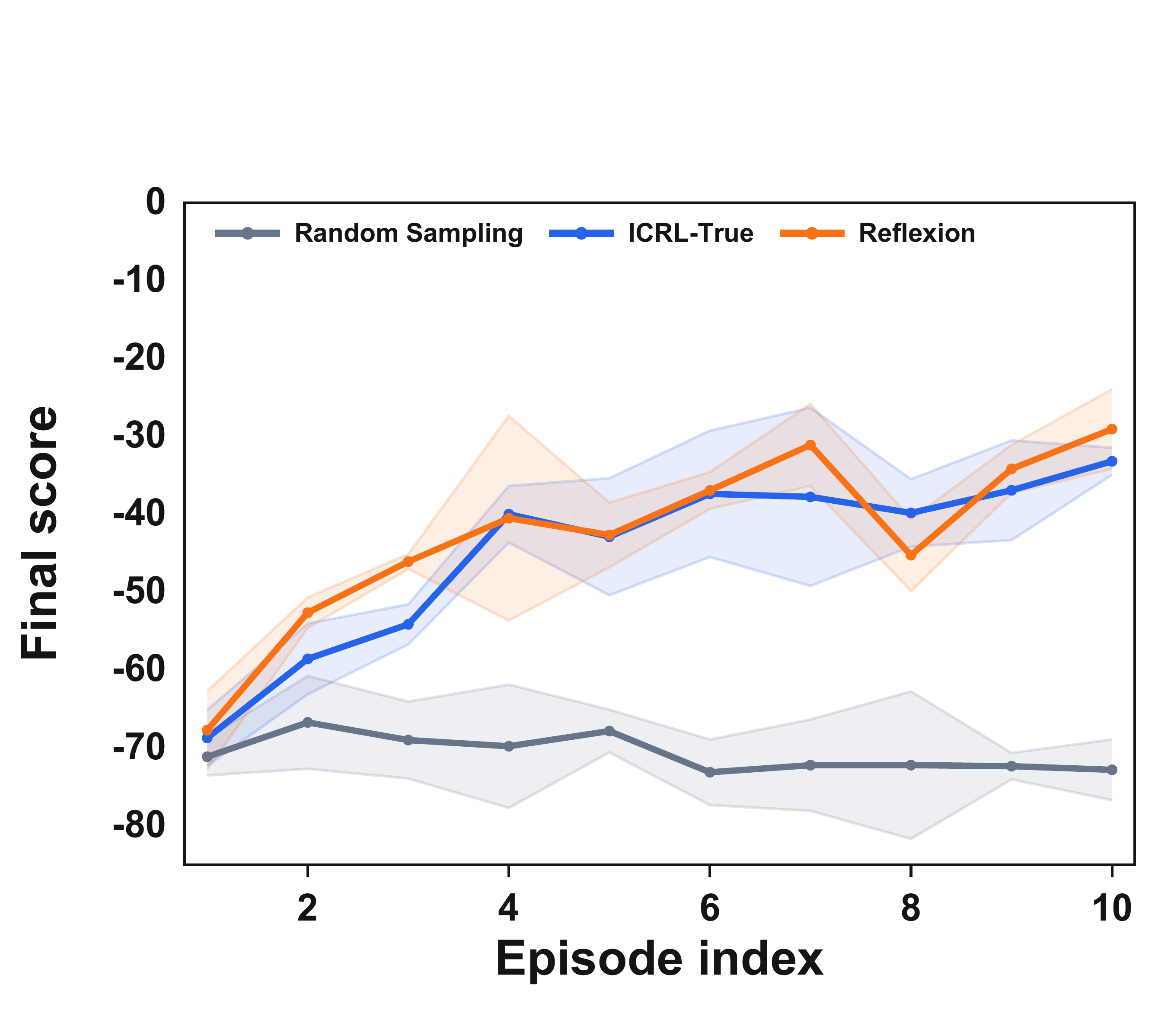}
    \caption{ScienceWorld ICRL experiment with Reflexion.}
    \label{fig:sw_reflexion}
\end{figure}




\begin{figure}[t]
    \centering
    
    \begin{subfigure}{0.45\textwidth}
        \centering
        \includegraphics[width=\linewidth]{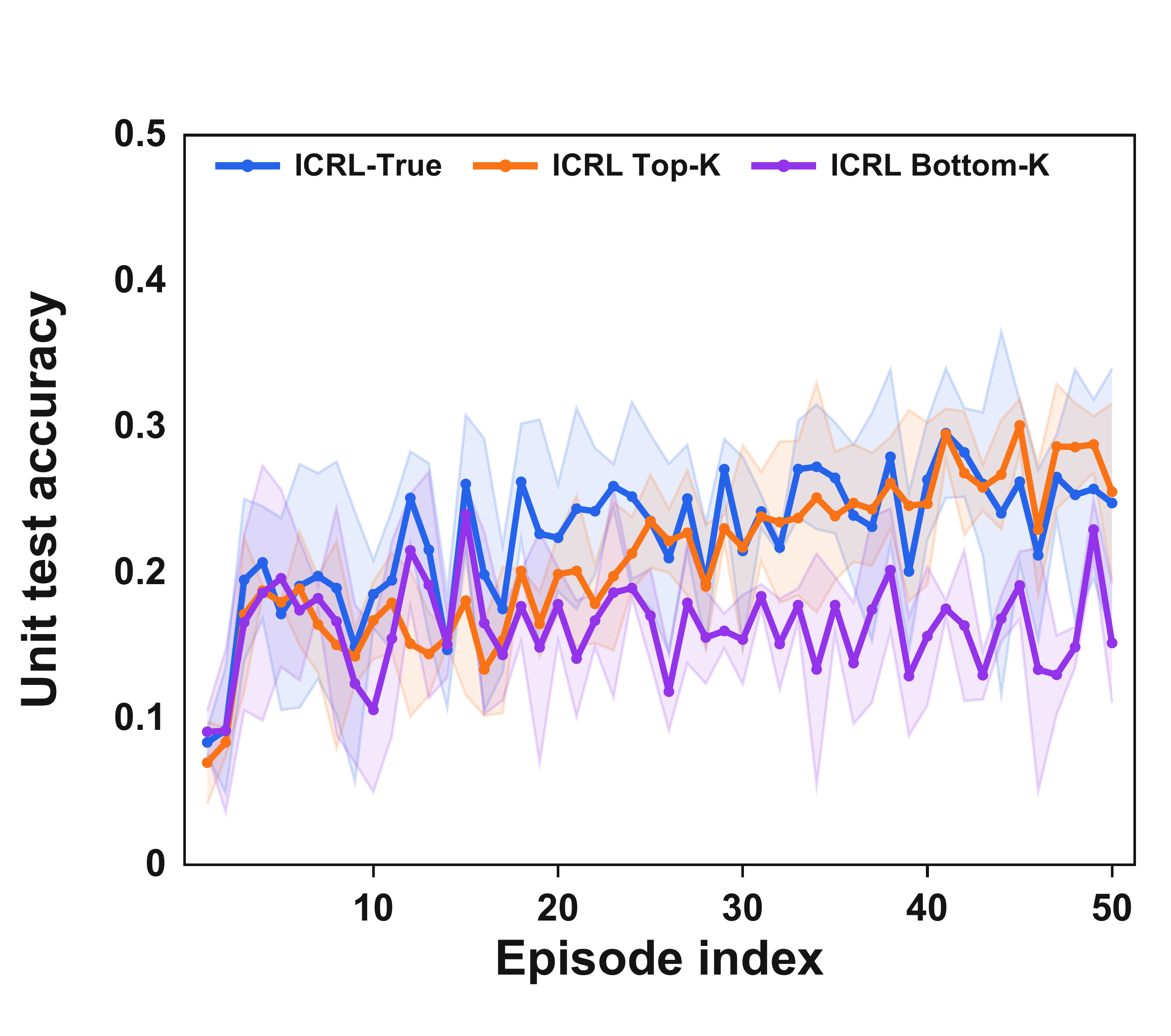}
        \caption{The topk and bottom k results of MBPP}
    \end{subfigure}
    \hfill
    \begin{subfigure}{0.45\textwidth}
        \centering
        \includegraphics[width=\linewidth]{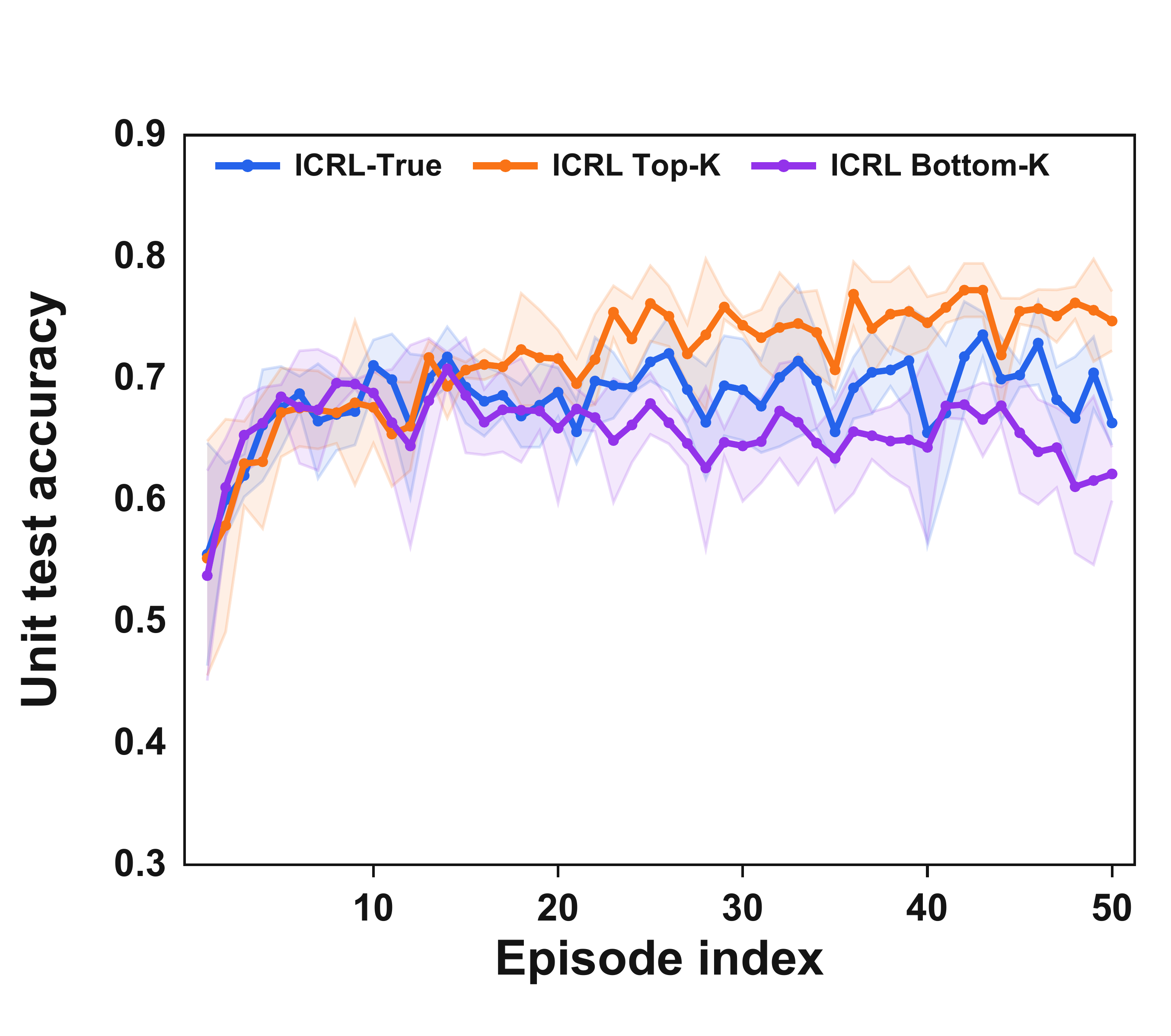}
        \caption{The topk and bottom k results of HumanEval}
    \end{subfigure}
    \hfill

    \caption{Top-k and Bottom-k memory selection on MBPP+ and HumanEval+.}
    \label{appen:topkbottomk}
\end{figure}
\paragraph{Memory selection}
In ~\Cref{appen:topkbottomk}, we compare groups that use Top-k selection, Bottom-k selection, and random selection as their memory policies.
The experimental results show that Top-k selection performs better than Bottom-k.
Random selection performs worse than Top-k selection under seed variance.

\subsection{Full Results of Section 5}
\subsubsection{Setup}
\label{subsec:full-setup}

We test how each piece of the in-context memory affects the model's output. The memory holds $N{=}10$ trajectories by default. Each trajectory contains a problem, a solution, and a reward. We change one part of the memory at a time and check how $\Delta{\text{LOO}}$ and the model's behavior change. All results are split by $\tau^-$ (the target is a failure trajectory) and $\tau^+$ (the target is a success trajectory).

\paragraph{ICRL variants.} We compare thirteen ways of building the memory.
\begin{itemize}[leftmargin=1.5em]
    \item \emph{True Reward.} The default setting. Each trajectory keeps its real reward.
    \item \emph{All Failure.} Every reward in the memory is set to $0$.
    \item \emph{Specific failure.} Only the target trajectory's reward is set to $0$; the other rewards stay the same.
    \item \emph{Flipped Reward.} Every reward in the memory is flipped ($r \leftarrow 1-r$).
    \item \emph{Specific flipped.} Only the target trajectory's reward is flipped.
    \item \emph{Random Reward.} Every reward is replaced with a random value.
    \item \emph{Verbal Reward.} Each reward becomes one of four words --- \emph{failure}, \emph{partial failure}, \emph{partial success}, or \emph{success} --- using four equal bins over the reward range.
    \item \emph{Anti prompt.} We add an instruction that tells the model to produce something different from what the memory shows.
    \item \emph{CoT.} The model first reads the memory and writes a short thought, and then produces its final answer.
    \item \emph{Order First} / \emph{Order Last}. The target trajectory is placed at the start or at the end of the memory.
    \item \emph{Duplicate.} We add five copies of the target trajectory, raising the memory size to $N{=}15$.
    \item \emph{Cross-Task.} The memory is filled with trajectories from a different task. We list it for completeness and do not analyze it because the trajectories share no structure with the target.
\end{itemize}

\paragraph{Reflexion variants.} \emph{Reflexion Only} shows the model only the abstract reflection text, without the raw trajectory. \emph{Reflexion + Trajectory} shows both the reflection text and the trajectory.

\paragraph{Trajectory Only.} The memory contains the trajectory but no reward.

\paragraph{Metrics.} $\Delta{\text{LOO}}$ is the leave-one-out effect on the score: we remove the target trajectory from the memory and measure how much the score drops. Target rate is the rate at which the model's output matches the target trajectory exactly. Memory rate is the rate at which the model's output matches any trajectory in the memory. Problem Only in ~\Cref{append:behavior_full} is a baseline that gives the model only the problem and no memory; its Memory Replay Rate (16.3\%) acts as a chance level for matching a held-out memory pool.

\subsubsection{Analysis}
\label{subsec:full-analysis}

\paragraph{Reward content has only a weak effect.} Across reward variants, $\Delta{\text{LOO}}$ moves by a small and inconsistent amount. True Reward gives $0.36 / 0.17$ ($\tau^- / \tau^+$). Flipped Reward gives $0.47 / 0.15$, All Failure gives $0.41 / 0.15$, Random Reward gives $0.42 / 0.15$, and Verbal Reward gives $0.40 / 0.15$. The numbers shift a little, but not in one clear direction. Memory rate in ~\Cref{append:behavior_full} tells the same story: it stays close to $30\%$ for every reward variant. We read this as weak evidence that the model uses the reward field, but the trend is not clean.

\paragraph{Target-only changes track the same pattern.} Changing only the target trajectory's reward (Specific failure, Specific flipped) gives $\Delta_{\text{LOO}}$ and behavior values close to the all-trajectory versions. The model does not react more strongly when the reward change is focused on the target.

\paragraph{Position and frequency matter more than reward.} Order Last gives a $\tau^-$ Target Replay Rate of $14.9\%$, far above True Reward at $3.0\%$. Duplicate is the highest at $22.8\%$. Both also push the $\tau^-$ Memory Replay Rate above $40\%$. The same effects are much weaker for $\tau^+$. \emph{Where} and \emph{how often} a trajectory appears in the memory has a larger effect than the reward attached to it.

\paragraph{Anti-prompt does not move behavior.} Adding an instruction that asks the model to do something different from the memory does not change $\Delta{\text{LOO}}$ ($0.33 / 0.16$) or Memory Replay Rate ($30.0 / 30.2\%$). The numbers stay close to True Reward. The instruction is not enough to pull the model away from the memory content.

\paragraph{Reflexion Only has a near-zero LOO effect.} Reflexion Only gives $\Delta{\text{LOO}}$ close to $0$ ($-0.006 / 0.005$), and its Memory Replay Rate ($\sim 11.7\%$) is below the Problem Only baseline. Without the trajectory in the memory, removing the target makes no real difference. Once the trajectory is added back (Reflexion + Trajectory: $0.44 / 0.14$), the numbers look like the other ICRL variants. Most of the LOO effect seems to come from the trajectory itself, not from the reflection.

\paragraph{Trajectory Only is close to ICRL.} Trajectory Only (no reward at all) gives $\Delta{\text{LOO}} = 0.42 / 0.16$ and Memory Replay Rate around $33\%$. This is very close to True Reward and to most ICRL variants. Dropping the reward field does not change the picture much, which lines up with the weak reward effects above.

\paragraph{Summary.} Reward changes give small and inconsistent effects, while changes to the trajectory itself --- its position, its count, or whether it is shown at all --- give larger effects. We read this as weak evidence that the model uses rewards in memory, but the dominant signal seems to be the trajectory content rather than the reward.
\begin{table}[]
\centering
\begin{tabular}{@{}llcc@{}}
\toprule
            &                         & \multicolumn{2}{c}{$\Delta{\mathrm{LOO}}$} \\ 
\cmidrule(l){3-4}
Method Type &                         & $\tau^-$ & $\tau^+$ \\ 
\midrule
ICRL        & True Reward             & 0.3601 & 0.1734 \\
            & All Failure             & 0.4066 & 0.1468 \\
            & Anti prompt             & 0.3322 & 0.1632 \\
            & CoT                     & 0.4339 & 0.1615 \\
            & Duplicate               & 0.4327 & 0.2504 \\
            &Specific flipped         & 0.4742 & 0.1459 \\
            & Order First             & 0.4823 & 0.2030 \\
            & Order Last              & 0.4656 & 0.2385 \\
            & Random Reward           & 0.4165 & 0.1528 \\
            & Specific failure        & 0.3601 & 0.1382 \\
            & Flipped Reward          & 0.4936 & 0.1512 \\
            & Verbal Reward           & 0.4037 & 0.1506 \\ 
\midrule
Reflexion   & Reflexion Only          & -0.0057 & 0.0051 \\
            & Reflexion + Trajectory  & 0.4445 & 0.1426 \\ 
\midrule
Trajectory Only & Trajectory Only     & 0.4159 & 0.1621 \\ 
\bottomrule
\end{tabular}
\caption{Full results of $\Delta{\mathrm{LOO}}$}
\label{appen:delta-loo}
\end{table}

\begin{table}[t]
\centering
\caption{Full replay-rate results across memory conditions.}
\label{append:behavior_full}
\begin{tabular}{llrrrr}
\toprule
\multirow{2}{*}{Type} & \multirow{2}{*}{Detail}
& \multicolumn{2}{c}{Target Replay Rate (\%)}
& \multicolumn{2}{c}{Memory Replay Rate (\%)} \\
\cmidrule(lr){3-4} \cmidrule(lr){5-6}
& & $\tau^+$ & $\tau^-$ & $\tau^+$ & $\tau^-$ \\
\midrule
Problem Only & Problem Only & 1.1 & 0.7 & 16.3 & 16.3 \\
\midrule
\multirow{13}{*}{ICRL}
& True Reward      & 3.0  & 2.5 & 30.5 & 30.5 \\
& All Failure      & 1.7  & 2.3 & 27.3 & 27.6 \\
& Specific Failure & 1.7  & 2.2 & 30.6 & 30.5 \\
& Anti Prompt      & 2.3  & 2.4 & 30.0 & 30.2 \\
& Cross-Task       & 0.0  & 4.7 & 13.3 & 13.4 \\
& Flipped Reward   & 1.1  & 3.1 & 29.4 & 29.6 \\
& Specific Flipped & 2.4  & 3.6 & 30.5 & 32.4 \\
& Random Reward    & 1.0  & 3.6 & 30.7 & 30.5 \\
& Verbal Reward    & 2.1  & 2.9 & 32.4 & 31.8 \\
& Order First      & 5.5  & 2.4 & 30.0 & 29.3 \\
& Order Last       & 14.9 & 3.2 & 41.9 & 29.4 \\
& Duplicate        & 22.8 & 3.5 & 45.4 & 32.1 \\
& CoT              & 5.7  & 1.3 & 28.6 & 28.2 \\
\midrule
\multirow{2}{*}{Reflexion}
& Reflexion Only         & 1.9 & 0.5 & 11.7 & 11.4 \\
& Reflexion + Trajectory & 3.0 & 1.4 & 18.1 & 18.2 \\
\midrule
Trajectory Only & Trajectory Only & 1.8 & 3.1 & 34.3 & 32.7 \\
\bottomrule
\end{tabular}
\end{table}

\subsection{Full Results of Section 6}
\label{appen:full_6}
\begin{table}[t]
\centering
\caption{Mean reward comparison across MBPP and ScienceWorld with standard deviations.}
\begin{tabular}{@{}lcc@{}}
\toprule
Memory Setting & MBPP & ScienceWorld \\ 
\midrule
RSM & $0.250 \pm 0.021$ & $-32.479 \pm 4.520$ \\
Cross-Task & $0.122 \pm 0.017$ & $-43.000 \pm 4.100$ \\
Random Valid Action & $0.102 \pm 0.012$ & $-68.208 \pm 3.880$ \\
ICRL History & $0.245 \pm 0.032$ & $-42.238 \pm 2.860$ \\
\midrule
\multicolumn{3}{@{}l@{}}{\textit{Trajectory Perturbation}} \\
\quad Shuffle & $0.243 \pm 0.023$ & $-38.588 \pm 6.090$ \\
\quad Crossover & $0.251 \pm 0.025$ & $-39.701 \pm 5.870$ \\
\quad Insert random token & $0.242 \pm 0.021$ & -- \\
\bottomrule
\end{tabular}%
\label{tab:full_6}
\end{table}
We present the complete results of the ~\Cref{tab:mean_reward_comparison} experiment, including the std, in ~\Cref{tab:full_6}.

\section{Qualitative Results}

\subsection{visualization of Crossover}
\begin{figure*}[t]
\centering

\begin{subfigure}[t]{0.48\textwidth}
\begin{lstlisting}[style=pycode]
def bell_Number(n):
    """Calculate the nth Bell number."""
    if n < 0:
        raise ValueError("...")
    if n == 0:
        return 1

    triangle = []
    triangle.append([1])

    for i in range(1, n + 1):
        prev_row = triangle[i - 1]
        current_row = [prev_row[-1]]
        for j in range(1, i + 1):
            current_row.append(
                current_row[j-1] + prev_row[j-1])
        triangle.append(current_row)
    return triangle[n][0]
\end{lstlisting}
\caption{Parent A}
\end{subfigure}
\hfill
\begin{subfigure}[t]{0.48\textwidth}
\begin{lstlisting}[style=pycode]
def bell_Number(n):
    """Calculate the nth Bell number.
    Args:
        n (int): The index.
    Returns:
        int: The nth Bell number.
    """
    if not isinstance(n, int) or n < 0:
        raise ValueError("...")
    if n == 0:
        return 1

    bell_triangle = [[0]*(n+1)
                     for _ in range(n+1)]
    bell_triangle[0][0] = 1
    for i in range(1, n + 1):
        bell_triangle[i][0] = (
            bell_triangle[i-1][i-1])
        for j in range(1, i + 1):
            bell_triangle[i][j] = (
                bell_triangle[i][j-1]
                + bell_triangle[i-1][j-1])
    return bell_triangle[n][0]
\end{lstlisting}
\caption{Parent B}
\end{subfigure}

\vspace{0.6em}

\begin{subfigure}{\textwidth}
\begin{lstlisting}[style=pycode,
  linebackgroundcolor={%
    \ifnum\value{lstnumber}=5 \color{parentBcol}\fi%
    \ifnum\value{lstnumber}=6 \color{parentBcol}\fi%
    \ifnum\value{lstnumber}=7 \color{parentBcol}\fi%
    \ifnum\value{lstnumber}=8 \color{parentBcol}\fi%
    \ifnum\value{lstnumber}=9 \color{parentBcol}\fi%
    \ifnum\value{lstnumber}=11 \color{parentBcol}\fi%
    \ifnum\value{lstnumber}=17 \color{parentBcol}\fi%
    \ifnum\value{lstnumber}=18 \color{parentAcol}\fi%
    \ifnum\value{lstnumber}=19 \color{parentBcol}\fi%
    \ifnum\value{lstnumber}=20 \color{parentAcol}\fi%
    \ifnum\value{lstnumber}=21 \color{parentBcol}\fi%
    \ifnum\value{lstnumber}=22 \color{parentAcol}\fi%
    \ifnum\value{lstnumber}=23 \color{parentBcol}\fi%
    \ifnum\value{lstnumber}=25 \color{parentBcol}\fi%
    \ifnum\value{lstnumber}=26 \color{parentAcol}\fi%
    \ifnum\value{lstnumber}=27 \color{parentBcol}\fi%
    \ifnum\value{lstnumber}=28 \color{parentAcol}\fi%
    \ifnum\value{lstnumber}=29 \color{parentAcol}\fi%
    \ifnum\value{lstnumber}=31 \color{parentAcol}\fi%
    \ifnum\value{lstnumber}=32 \color{parentAcol}\fi%
    \ifnum\value{lstnumber}=33 \color{parentAcol}\fi%
    \ifnum\value{lstnumber}=34 \color{parentAcol}\fi%
    \ifnum\value{lstnumber}=36 \color{parentAcol}\fi%
    \ifnum\value{lstnumber}=38 \color{parentAcol}\fi%
    \ifnum\value{lstnumber}=39 \color{parentAcol}\fi%
  }]
def bell_Number(n):
    """
    Calculate the nth Bell number.

    Args:
        n (int): The index of the Bell number to calculate.

    Returns:
        int: The nth Bell number.
    """
    if not isinstance(n, int) or n < 0:
        raise ValueError("Input must be a non-negative integer.")

    if n == 0:
        return 1

    # Initialize Bell triangle
    # We need rows 0 to n. Each row i has i+1 elements.
    bell_triangle[0][0] = 1
    triangle = []
    # Fill the triangle
    # Row 0
        bell_triangle[i][0] = bell_triangle[i - 1][i - 1]

            bell_triangle[i][j] = bell_triangle[i][j - 1] + bell_triangle[i - 1][j - 1]
    for i in range(1, n + 1):
    return bell_triangle[n][0]
        prev_row = triangle[i - 1]
        current_row = [prev_row[-1]]

        # Fill the rest of the row
        for j in range(1, i + 1):
            # Element is sum of left and above
            current_row.append(current_row[j - 1] + prev_row[j - 1])

        triangle.append(current_row)

    # The nth Bell number is the first element of the nth row
    return triangle[n][0]
\end{lstlisting}
\caption{Child: lines from \colorbox{parentAcol}{\strut\,Parent A\,} and 
\colorbox{parentBcol}{\strut\,Parent B\,} are alternated. 
\textbf{This code does not run}: \texttt{bell\_triangle} is referenced before being defined, 
and the body of the outer \texttt{for} loop is split across the \texttt{return} statement. 
Indentation is preserved exactly as produced.}
\end{subfigure}

\caption{Example of the \textbf{Crossover} perturbation on an MBPP+ task. 
Two parent trajectories (a, b), each a valid solution sampled by the model, 
are mixed line-by-line into a structurally broken child (c). 
Even though the child code does not compile, 
using such perturbed trajectories as in-context memory yields performance 
comparable to using real ICRL trajectories 
(~\Cref{sec:exp3}, ~\Cref{tab:full_6}).}
\label{fig:crossover-example}
\end{figure*}

\begin{figure}[t]
\centering
\begin{subfigure}[t]{0.46\textwidth}
\centering
\begin{lstlisting}[
  basicstyle=\ttfamily\small,
  frame=single,
  numbers=left,
  numberstyle=\color{gray}\small,
  framesep=4pt,
  xleftmargin=2em,
  breaklines=true,
  breakindent=0pt,
  postbreak=\mbox{\textcolor{gray}{$\hookrightarrow$}\space},
]
focus on potato
focus on red box
move orange in inventory to red box
focus on potato
move potato to red box
\end{lstlisting}
\caption{Original trajectory}
\end{subfigure}
\hfill
\begin{subfigure}[t]{0.46\textwidth}
\centering
\begin{lstlisting}[
  basicstyle=\ttfamily\small,
  frame=single,
  numbers=left,
  numberstyle=\color{gray}\small,
  framesep=4pt,
  xleftmargin=2em,
  breaklines=true,
  breakindent=0pt,
  postbreak=\mbox{\textcolor{gray}{$\hookrightarrow$}\space},
]
focus on potato
focus on potato
focus on red box
move orange in inventory to red box
move potato to red box
\end{lstlisting}
\caption{Shuffled trajectory}
\end{subfigure}
\caption{Example of the Shuffle perturbation on a ScienceWorld task. The first and last actions are kept fixed; the middle three steps are randomly reordered. The shuffled version is no longer a sensible plan --- \texttt{focus on potato} is repeated immediately at step 2, and the agent moves the potato to the red box at step 5 without re-focusing on it --- yet using such trajectories as in-context memory yields performance comparable to using the original trajectories (\Cref{sec:exp3}).}
\label{fig:shuffle_example}
\end{figure}

We show an actual crossover sample in ~\Cref{fig:crossover-example}.
Between 1 and 3 lines are selected at random, and there is a 50\% chance of determining which parent they will inherit from.
In practice, most of the code does not compile and results in syntax errors.

\subsection{Visualization of Shuffle}
\label{app:shuffle_qual}

We show an actual Shuffle sample in \Cref{fig:shuffle_example}. 
The first and last actions are kept fixed, and the middle steps are randomly reordered. 
Although the resulting trajectory is no longer a coherent plan --- the agent repeats the same focus action or attempts to move an object without focusing on it first --- using such trajectories as in-context memory yields performance comparable to using the original trajectories (\Cref{sec:exp3}). 
This supports our claim in \Cref{sec:exp3} that the model benefits from the surface form of the trajectory rather than its causal or semantic content.

\newpage

\end{document}